\documentclass{article}
\usepackage{amsmath, amssymb}
\usepackage{graphicx}
\usepackage[export]{adjustbox}
\usepackage{caption}
\usepackage{subcaption}
\usepackage{makecell}
\usepackage{adjustbox} 
\usepackage{geometry}
\usepackage{authblk}
\usepackage{cite}
\usepackage{pgfplots}
\pgfplotsset{compat=1.18} 
\usepackage{float} 
\usepackage{algorithm}
\usepackage{algorithmic}
 \usepackage{tikz}
 \usetikzlibrary{positioning,arrows.meta}
\usepackage{float}
\usepackage{hyperref}
\usepackage{multirow}
\usepackage{booktabs,tabularx}
\title{Single Image Defogging Using a Fourth-Order Telegraph PDE Guided by Physical Haze Modeling
}
 \author{
    \textbf{Manish Kumar$^{1*}$,Rajendra K. Ray $^{2}$} \\
    \textsuperscript{1,2} School of Mathematical and Statistical Sciences, \\
    Indian Institute of Technology Mandi, Mandi (H.P), 175075, India \\
    \texttt{manishkumarkitlana@gmail.com}, \texttt{rajendra@iitmandi.ac.in} \\
    \textsuperscript{*}\footnote{Corresponding author: Rajendra K. Ray, \texttt{rajendra@iitmandi.ac.in}}}

\begin{document}
\maketitle

\begin{abstract}

In real-world scenarios, image defogging is an inverse problem due to unknown scene depth, atmospheric scattering, and the common absence of ground truth . To resolve the issue, we propose a hybrid defogging model that integrates a fourth-order nonlinear PDE with a physical haze formation model. We used Dark Channel Prior to estimate atmospheric parameters and to generate a guidance image, while the final restoration is performed via a fourth-order PDE-based evolution. A fourth-order PDE of the type telegraph is then evolved, incorporating an edge-adaptive diffusion coefficient and a fidelity term weighted by the transmission map. Fourth-order diffusion effectively suppresses haze while preserving structural details, and the hyperbolic formulation improves numerical stability and convergence behavior. We use relative error norm criteria for the convergence of our PDE. The proposed method is compared with Dark Channel prior, modified Dark Channel prior, and variational-based single-image defogging techniques. When we have ground truth available, we use MSE and SSIM for quantitative evaluation, whereas no-reference metrics, including FADE, Contrast Restoration Index, Average Gradient, and Entropy, are applied to real-world foggy images. Experimental results demonstrate that the proposed hybrid PDE-based method provides comparable visual quality and maintains structural details.
\end{abstract}

\section{Introduction}

Outdoor vision scenario are highly subject to degradation when operating under bad atmospheric conditions, such as fog and haze. These weather conditions arise because of suspension of fine particles in the atmosphere, which interact with light as it travels from objects in the scene to the camera sensor. In this way, photons experience repeated scattering and absorption events, causing a reduction of the original scene radiance and the introduction of airlight. As a result, the captured images losses contrast, color fidelity, and fine structural details, and severe visibility degradation, particularly for objects located at depths \cite{narasimhan2002vision, narasimhan2003contrast}. Such kind of degradation complicates image analysis.
 For modern intelligent systems atmospheric degradation poses crucial challenges. Applications such as autonomous driving, intelligent transportation systems, traffic surveillance, and satellite remote sensing depends on accurate visual perception. The presence of fog disturb image features such as edges, textures, and gradients, which is crucial for higher-level tasks, such as object detection, tracking, and scene understanding. Consequently, reduced visual clarity can induce cascading failures in subsequent computer vision modules, which may in turn result in erroneous decision-making and elevated safety risks in real-world deployment scenarios\cite{pandey2025comprehensive}. Therefore, robust defogging algorithms are crucial as preprocessing modules for outdoor vision systems.
The formation of foggy images is commonly described using the Atmospheric Scattering Model, originally proposed by Koschmieder \cite{koschmieder1924theorie} and modified for computer vision applications. 
The formation of foggy images is defined as:
\begin{equation}
    u(x) = J(x)T(x) + A(1 - T(x))
\end{equation}
Here, $u(x)$ denotes the observed intensity of the hazy input at pixel $x$, while $J(x)$ represents the haze-free scene radiance to be recovered. The parameter $A$ refers to the global atmospheric light, and $T(x)$ is the transmission map, which quantifies the portion of light that reaches the camera sensor without undergoing scattering.
The transmission function follows the Beer-Lambert law \cite{mccartney1976optics} and depends exponentially on scene depth and atmospheric scattering density. As a result, distant objects are more severely Suppressed and appear increasingly with atmospheric light, resulting in the characteristic depth-dependent haze effect.
Recovering the haze-free scene radiance $J(x)$ from a single hazy image is a highly ill-posed inverse problem. At each pixel, only one measurement is observed, yet several quantities—the transmission map and at the same time the atmospheric light—must be inferred. Early methods attempted to resolve this Uncertainty by using multiple images of the same scene taken under different weather conditions.\cite{narasimhan2003contrast} or by using polarization-based imaging systems \cite{schechner2001instant}. Although these methods provide accurate estimates, they require controlled acquisition conditions, static scenes, or specialized hardware, making them unsuitable for real-time and dynamic scenarios such as autonomous surveillance. To overcome these limitations, modern research has increasingly focused on single-image defogging techniques, which aim to recover scene radiance from a single degraded image by leveraging statistical priors, heuristic assumptions, or learned representations \cite{tan2008visibility}. Among these, the Dark Channel Prior (DCP) \cite{he2009single} has played a pivotal role by providing an effective statistical observation for estimating transmission maps in natural outdoor scenes. In parallel, learning-driven methods, particularly convolutional neural networks, have achieved encouraging performance by directly learning complex nonlinear transformations from hazy inputs to clean images \cite{cai2016dehazenet}. However, despite these advances, such approaches still exhibit drawbacks, including halo artifacts, color distortions in high-intensity areas, sensitivity to parameter selection, and poor generalization when confronted with real-world haze conditions that deviate from the synthetic scenarios used during training \cite{zhu2015fast}.

\section{Literature Review}

Single-image generation has been an active research topic over the past decade, leading to the development of a wide range of techniques based on different theoretical foundations. Existing methods can generally be divided into three groups: image enhancement–based approaches, physical model–based (prior-driven) approaches, and learning-based approaches. Each group has its own characteristic advantages and drawbacks with respect to visual quality, computational cost, and robustness.
Early image enhancement-based methods treat defogging primarily as a contrast enhancement problem, without explicitly modeling the physical process of atmospheric scattering. Classical approaches such as Histogram Equalization and Contrast Limited Adaptive Histogram Equalization (CLAHE) \cite{pizer1987adaptive} enhance visual contrast by redistributing image intensity values. Although these techniques are computationally efficient and straightforward to apply, they tend to amplify noise in homogeneous regions like the sky and do not adequately recover depth-dependent visibility, which can lead to visually unrealistic results \cite{tan2008visibility}. Retinex-based methods \cite{land1977retinex}, which attempt to decompose an image into illumination and reflectance components, can improve local contrast but often produce halo artifacts and color distortions because they struggle to distinguish haze-related effects from  illumination changes.
Physical model-based approaches constitute the mainstream of defogging research by explicitly leveraging the atmospheric scattering model. Among these, the Dark Channel Prior (DCP) proposed by He et al. \cite{he2009single, he2011single} remains one of the most influential methods. By exploiting the observation that at least one color channel exhibits very low intensity in haze-free outdoor patches, DCP enables effective estimation of transmission maps in a wide range of natural scenes. However, its underlying assumptions break down in bright regions such as the sky or white objects, leading to over-saturation and color distortions. Moreover, the initial block-wise transmission estimates require refinement, and the original soft matting approach \cite{levin2008closed} incurs significant computational overhead. Among these, Salazar-Colores \emph{et al.}
introduced a fast image dehazing algorithm based on morphological reconstruction, which significantly
reduces computational complexity while preserving the core physical assumptions of the atmospheric
scattering model \cite{salazar2019fast}.
To improve computational efficiency, guided filtering was introduced as a fast alternative for transmission refinement \cite{he2013guided}. Guided filtering preserves strong edges while achieving linear-time complexity, making it suitable for practical applications. Nevertheless, it remains sensitive to parameter selection and may produce halo artifacts near strong depth discontinuities. To further enhance performance, several studies explored alternative color spaces and fusion strategies. For example, Tufail et al. \cite{tufail2018improved} proposed fusing transmission estimates obtained from RGB and YCbCr color spaces, exploiting the separation of luminance and chrominance components. Although this strategy improves structural detail preservation, it increases computational complexity and introduces additional fusion parameters.
Segmentation-based methods have been proposed to explicitly address the failure of prior-based approaches in sky regions. Sabir et al. \cite{sabir2020segmentation} and Anan et al. \cite{anan2021image} introduced sky and non-sky segmentation to apply different atmospheric light estimation strategies. While these methods effectively reduce over-enhancement in bright regions, their performance heavily depends on accurate segmentation. Errors at region boundaries often result in visible seams or artifacts, and the segmentation process adds further computational overhead.
Real-time defogging has also received considerable attention, particularly for video surveillance and autonomous driving applications. Wu et al. \cite{wu2024real} proposed an adaptive threshold-based framework for high-definition video defogging, while Li et al. \cite{li2017improved} employed quad-tree subdivision and gain coefficients to replace expensive matting operations. Although these approaches achieve real-time performance, they often sacrifice fine texture details due to algorithmic approximations.
Depth-aware and geometry-based approaches attempt to incorporate three-dimensional scene information into defogging. Methods proposed by Kokul and Anparasy \cite{kokul2020single} and Gibson and Nguyen \cite{gibson2013analysis} estimate coarse depth cues or analyze color distributions to improve transmission estimation. Despite their conceptual advantages, depth estimation from a single image remains fundamentally ambiguous, and inaccuracies can propagate into the defogging process, particularly in textured regions that mimic depth variations.
More recently, learning-based methods using convolutional neural networks have demonstrated impressive performance on benchmark datasets. DehazeNet \cite{cai2016dehazenet} and AOD-Net \cite{li2017aod} learn end-to-end mappings from hazy inputs to transmission maps or restored images, achieving high PSNR and SSIM values. However, these methods rely heavily on synthetic training data and often suffer from domain shift when applied to real-world scenes. Additionally, their dependence on high-performance computing resources limits their applicability in resource-constrained environments.
Despite the diversity of existing approaches, a persistent challenge remains in balancing noise suppression and edge preservation. Excessive smoothing removes haze but blurs structural boundaries, while insufficient smoothing leaves residual artifacts. Recently, Majee et al.~\cite{majee2020despeckling} introduced a telegraph diffusion equation (TDE)–based approach for mitigating speckle noise in images. Although the method demonstrates effective noise suppression, the use of second-order diffusion often leads to blocky or staircase artifacts in homogeneous regions. To address this limitation, You and Kaveh~\cite{you2000fourth} proposed a fourth-order partial differential equation model that significantly alleviates blocky effects by promoting smoother intensity transitions. However, pure fourth-order diffusion is known to amplify speckle-like oscillations and may oversmooth fine structures.

Image dehazing is often treated as an energy minimization problem using partial differential equations (PDE) and variational methods. Early works, such as Tarel and Hautiere \cite{tarel2009fast}, utilized median filtering within a PDE framework to estimate the atmospheric veil while preserving edges. Following this, more advanced variational models were developed. Fang et al. \cite{fang2014single} proposed a method that performs dehazing and denoising simultaneously, using a weighted total variation regularizer to prevent noise amplification during the dehazing process. Similarly, Galdran et al. \cite{galdran2015enhanced} focused on contrast enhancement, formulating an energy function that maximizes local contrast without explicitly estimating the transmission map depth.

Liu et al. \cite{liu2019unified} introduced a unified variational model that utilizes Dark Channel Prior (DCP) with Total Variation (TV) regularization to refine the transmission map. This model successfully reduces artifacts common in standard DCP. Specifically, first-order regularization tends to produce the staircase effect. Also, the fidelity term is strictly constrained to the initial dark channel, which gives color distortion in bright white objects.
Motivated by these observations, several studies have explored fourth-order PDE-based image denoising frameworks with adaptive diffusion control. In particular, Ray et al.~\cite{ray2025new} presented a fourth-order despeckling model incorporating a grayscale-based diffusion indicator, which effectively balances speckle suppression and edge preservation. Their approach demonstrates that properly designed adaptive fourth-order diffusion can reduce speckle noise while avoiding blocky artifacts. Most existing single-image defogging methods, such as Dark Channel Prior (DCP), aim to enhance visibility from a single hazy image. They often introduce halo effects, color distortions, and performance degradation in scenes with dense haze.
 Motivated by .~\cite{ray2025new} \cite{he2009single, he2011single} \cite{salazar2019fast}\cite{liu2019unified}, this paper proposes a Partial Differential Equation (PDE)-based framework for single image defogging that aims to suppress fog while preserving sharp structural details.
 
\section{Proposed Method}

Fog and haze generally degrade visibility by lowering contrast and fine details. To address this, We proposed a method that improves image clarity by integrating a physical model of haze formation with a fourth-order nonlinear partial differential equation (PDE). This model treats a foggy image as a superposition of the true scene radiance and the atmospheric light. For a image, the relationship at pixel $x$ is defined as:

\begin{equation}
    u(x) = J(x)\, T(x) + A\,(1 - T(x)),
\end{equation}
where $u(x)$ is the observed image degraded by fog. $J(x)$ is the latent radiance of the haze-free scene.  $A$ represents the global atmospheric light. $T(x)$ is the transmission map, indicating the percentage of light reaching the camera without scattering.
This work utilizes the statistical properties associated with the Dark Channel Prior. This principle suggests that in haze-free outdoor images, local patches typically contain at least one color channel with very low intensity. Mathematically, the dark channel $u_{\text{dark}}(x)$ is calculated as:

\begin{equation}
    u_{\text{dark}}(x) = \min_{y \in \Omega(x)} \left( \min_{c \in \{r,g,b\}} u^c(y) \right),
\end{equation}
where $\Omega(x)$ is a local patch centered at $x$, and $u^c$ is the intensity of color channel $c$. In the presence of haze, this value increases due to the additive airlight.
To estimate the atmospheric light $A$, we identify the brightest pixels within the dark channel $u_{\text{dark}}$, as these represent the most hazy regions. We select a small percentage of these pixels and determine the maximum intensity at the corresponding coordinates in the original image $u$.
Using this estimate, we calculate a preliminary transmission map. To prevent the complete removal of depth cues, a constant factor $\omega$ (where $0 < \omega < 1$) is introduced:

\begin{equation}
    t_{\text{rough}}(x) = 1 - \omega \frac{u_{\text{dark}}(x)}{A}.
\end{equation}

Patch-based calculation can introduce block artifacts, so this rough map is refined using a Gaussian smoothing filter $G_\sigma$.
Finally, the initial haze-free image is recovered by inverting the scattering model. To ensure numerical stability, a lower bound $T_0$ is applied to the transmission values.

\begin{equation}
    J(x) = \frac{u(x) - A}{\max(T(x),\, T_0)} + A.
\end{equation}
In our proposed method, this reconstructed $J(x)$ serves as a structural guide rather than the final result. It acts as a guidance image for a fourth-order PDE evolution.
To refine the guidance image and remove fog, we evolve the following telegraph fourth-order PDE:
\begin{equation}
u_{tt} + \lambda u_t
=
-\Delta \big( g(u)\,\Delta u \big)
- \lambda\, T^2(x)\,(u - J),
\end{equation}
where \(u(x,t)\) is the evolving image and \(\lambda\) controls the strength of the fidelity term.
The adaptive diffusion coefficient in this work is defined as
\begin{equation}
g(u_\xi, |\Delta u_\xi|) =
\frac{2 |u_\xi|^\alpha}{M_\xi^\alpha + |u_\xi|^\alpha}
\cdot
\frac{1}{1 + \left(\frac{|\Delta u_\xi|}{k}\right)^2},
\end{equation}
where $u_\xi$ denotes the smoothed image intensity, $\Delta u_\xi$ represents its Laplacian, and
\[
M_\xi = \max_{x \in \Omega} |u_\xi(x,t)|
\]
is the maximum intensity value over the image domain $\Omega$. The parameters $\alpha$ and $k$ control the sensitivity of the diffusion process.
The diffusion coefficient regulates smoothing based on local intensity information. It assigns a lower diffusion strength to regions with structures, thereby preventing visually important features. The Laplacian magnitude acts as a curvature term, which suppresses diffusion in areas where there are strong spatial variations.
To ensure numerical stability and robustness against fog, the image is preprocessed using Gaussian smoothing expressed as
\[
u_\xi = J_\xi \ast u,
\]
where $J_\xi$ denotes a Gaussian kernel.

\section{Numerical Discretization}

Given a hazy RGB image $u = (u^R, u^G, u^B)$, the dark channel is computed as
\begin{equation}
u^{\text{dark}}(x) = \min_{y \in \Omega(x)} \left( \min_{c \in \{R,G,B\}} u^c(y) \right),
\end{equation}
where $\Omega(x)$ denotes a local patch centered at pixel $x$. The atmospheric light $A$ is estimated as the highest percentile value of the dark channel.
The transmission map is then approximated by T(x).
Let $u^n_{i,j,c}$ denote the estimated image intensity at spatial location $(i,j)$, color channel $c \in \{R,G,B\}$, and discrete time level $n$.

The spatial derivatives are computed with second-order central finite-difference schemes on a uniform grid. For each color channel, the discrete gradient components at time level $n$ are given by
\begin{equation}
\nabla_x u^n_{i,j}
\approx
\frac{u^n_{i+1,j} - u^n_{i-1,j}}{2\,\delta x},
\qquad
\nabla_y u^n_{i,j}
\approx
\frac{u^n_{i,j+1} - u^n_{i,j-1}}{2\,\delta y},
\end{equation}
where $\delta x$ and $\delta y$ denote the spatial step sizes.
The Laplacian operator is discretized as
\begin{equation}
\Delta u^n_{i,j}
\approx
\frac{u^n_{i+1,j} - 2u^n_{i,j} + u^n_{i-1,j}}{(\delta x)^2}
+
\frac{u^n_{i,j+1} - 2u^n_{i,j} + u^n_{i,j-1}}{(\delta y)^2}.
\end{equation}
The temporal derivatives are discretized using a second-order finite difference scheme. Specifically, the second-order temporal derivative is approximated by
\begin{equation}
u_{tt} \approx \frac{u^{n+1} - 2u^n + u^{n-1}}{(\delta t)^2},
\end{equation}
while the first-order temporal derivative is approximated as
\begin{equation}
u_t \approx \frac{u^{n+1} - u^n}{\delta t},
\end{equation}
where $\delta t$ denotes the temporal step size.
Substituting these approximations into the equation and rearranging terms yields the explicit update rule
\begin{equation}
u^{n+1}_{i,j,c}
=
\frac{
(2 + \lambda \delta t) u^n_{i,j,c}
- u^{n-1}_{i,j,c}
- (\delta t)^2
\left[
\Delta \left( C^n_{i,j,c} \, \Delta u^n_{i,j,c} \right)
+ \lambda_f \, T^2_{i,j} \, (u^n_{i,j,c} - J_{i,j,c})
\right]
}
{1 + \lambda \delta t}.
\end{equation}
The scheme is iterated until convergence.
A convergence criterion is required to ensure numerical stability and to terminate the evolution once a steady state is reached. In this work, convergence is monitored using the relative error.
Let $u^n$ and $u^{n+1}$ denote the restored color image at two consecutive time levels $n$ and $n+1$, respectively. The relative error is defined as
\begin{equation}
\text{RelErr}^{\,n}
=
\frac{\| u^{n+1} - u^n \|_2}{\| u^n \|_2 + \varepsilon},
\end{equation}
where $\|\cdot\|_2$ denotes the discrete $\ell_2$ norm computed over all pixels and color channels, and $\varepsilon$ is a small positive constant introduced to avoid division by zero. A decreasing relative error indicates that the numerical solution is approaching a steady state.
In practice, the iteration is terminated when the relative error falls below a prescribed tolerance toll, i.e.,
\begin{equation}
\text{RelErr}^{\,n} < toll,
\end{equation}

\section{Experiments and Performance parameters}
\begin{figure}[H]
    \centering
    \begin{tabular}{cccccc}
        \includegraphics[width=0.15\textwidth]{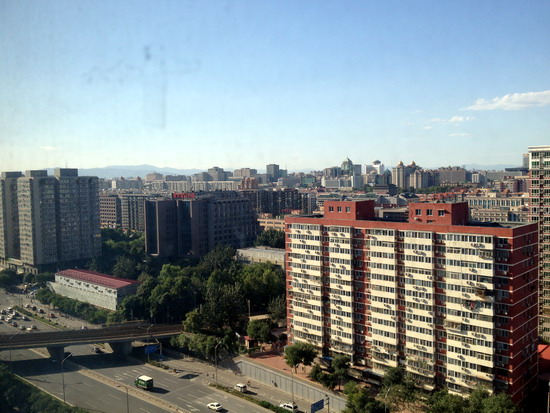} &
        \includegraphics[width=0.15\textwidth]{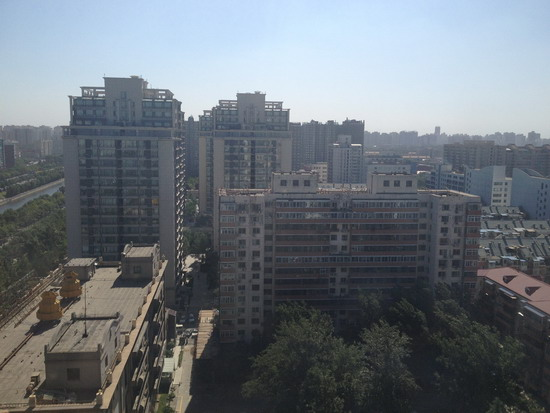} &
        \includegraphics[width=0.15\textwidth]{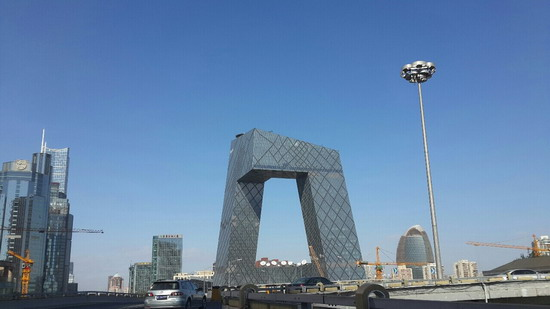} &
        \includegraphics[width=0.15\textwidth]{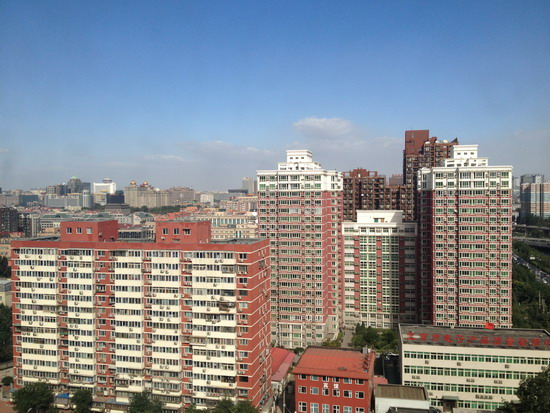} &
        \includegraphics[width=0.15\textwidth]{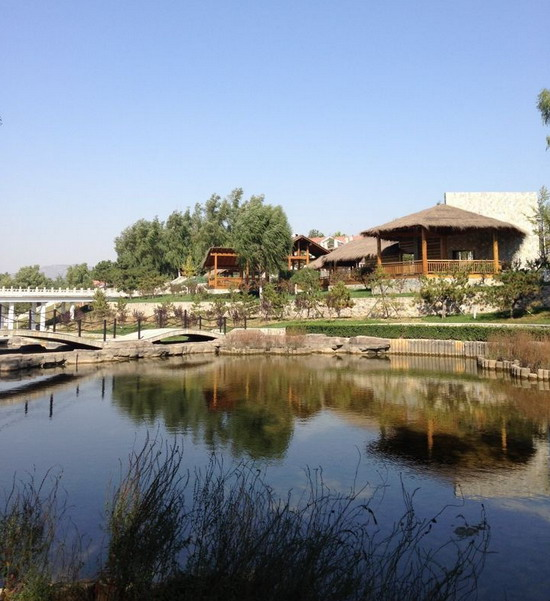} &
        \includegraphics[width=0.15\textwidth]{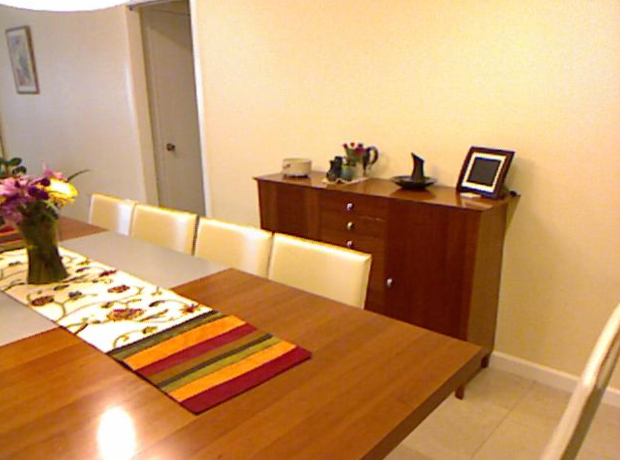} \\
        (a) Img 1 & (b) Img 2 & (c) Img 3 & (d) Img 4 & (e) Img 5 & (f) Img 6
    \end{tabular}
    \caption{Ground truth images}
    \label{fig:image1}
\end{figure}

This section presents a comprehensive evaluation of the proposed fourth-order telegraph PDE-based defogging model and compares its performance with three widely used prior-based approaches, namely dark channel prior (DCP)\cite{he2011single}, modified dark channel prior (MDCP) \cite{salazar2019fast}, and variational based SIDVBM\cite{liu2019unified}. We use six color images for conducting experiments with ground truth, each degraded by fog at three different levels, namely 10\%, 20\%, and 30\%. Varying fog densities allow a assessment of robustness under heavy degradation conditions.
All numerical experiments were implemented in MATLAB and executed on a system equipped with an Intel Core i7 processor running at 3.60~GHz and 8~GB of RAM. The proposed method uses an explicit finite-difference scheme for solving the fourth-order telegraph PDE. To guarantee numerical stability, the time step was chosen in accordance with the Courant--Friedrichs--Lewy (CFL) condition \cite{zauderer2011partial,li2009numerical,zhang2014tensor}. Specifically, the stability constraint is given by
\begin{equation}
\tau \leq \frac{h}{\max g(x,t)},
\end{equation}
where $h$ denotes the spatial grid size and $g(x,t)$ represents the adaptive diffusion coefficient. Based on this condition, a time step of $\tau = 0.05$ is chosen for our proposed model and existing variational based model, while the gaussian convolution parameter was fixed at $\xi = 2$.
The iterative evolution was terminated using a relative error-based convergence criterion. The stopping tolerance value is $10^{-4}$ for all test images and fog levels. This stopping criteria ensures consistent convergence behavior across experiments, In all experiments, the parameters of the proposed method are set as \( v = 1 \), \( \lambda = 1.5 \), and \( k = 2 \). The atmospheric light control parameter \( \omega \) is fixed at \(0.95\) for both the proposed method and the Dark Channel Prior (DCP) approach, And \( \omega = 0.75 \) is used for the Modified Dark Channel Prior (MDCP) method.
Figure~\ref{fig:image1} shows six original ground truth images. 
In next section we shows visual inspection and quantitative metrics demonstrate that the proposed model effectively suppresses haze while preserving fine details and avoiding halo and blocky artifacts commonly observed in prior-based methods.
\section{Image Quality Measurement}

In this section we presents a quantitative and qualitative evaluation of the proposed PDE-based defogging model with respect to fog reduction. To ensure a fair camparision, all experiments are carried out on a different set of color images illustrated in Fig.~\ref{fig:image1}.
Specifically, Images~1, 2, and~4 in Fig.~\ref{fig:image1} have spatial dimensions of $620 \times 413$ pixels, while Image~3 has a resolution of $550 \times 309$ pixels. Image~5 is of size $550 \times 601$ pixels, and Image~6 has a resolution of $620 \times 460$ pixels.
All test images are degraded by synthetic fog of varying densities. The restored results obtained by proposed approach are compared with classical Dark Channel Prior (DCP), modified variant (MDCP) and variational based approach (SIDVBM). The quality of the defogged images is check by using image quality metrics that capture both accuracy and structural information.
In this work, the mean squared Error(MSE) and the structural similarity index measure (SSIM) are evaluate when the ground truth image is available.
In addition to numerical measures, visual comparisons are provided to highlight improvements in contrast enhancement, edge sharpness, and color consistency achieved by the proposed model.

\subsection{Mean Squared Error (MSE)}

MSE measures the average variance at the pixel-level found when comparing the defogged result with the ground truth.
Let $u(i,j)$ denote the ground truth image and $\hat{u}(i,j)$ represent the
restored image, both defined on an image domain of size $M \times N$.
The MSE is defined as
\begin{equation}
\text{MSE} =
\frac{1}{MN}
\sum_{i=1}^{M}
\sum_{j=1}^{N}
\left( u(i,j) - \hat{u}(i,j) \right)^2.
\end{equation}
MSE directly measures the energy of the reconstruction error.
A lower value of MSE indicates that the defogged image is numerically closer
to the ground truth, implying more accurate recovery of intensity values.
Although MSE does not explicitly model human visual perception, it provides
a clear and objective indication of reconstruction fidelity and is therefore
widely used for algorithmic comparison.

\subsection{Structural Similarity Index Measure (SSIM)}

SSIM tell us the similarity between two images by comparing their structural patterns rather than relying solely on pixel-wise errors.
For image patches $i$ and $j$ extracted from the reference and restored images,
The mathematical formulation is:

\begin{equation}
\text{SSIM}(i,j) =
\frac{(2\mu_i\mu_j + C_1)(2\sigma_{ij} + C_2)}
{(\mu_i^2 + \mu_j^2 + C_1)(\sigma_i^2 + \sigma_j^2 + C_2)}.
\end{equation}
Here, $\mu$ represents the average brightness (mean), while $\sigma^2$ represents the contrast (variance) of the respective image windows. The term $\sigma_{ij}$ signifies the covariance. To prevent numerical errors when the denominator is near zero, small parameters $C_1$ and $C_2$ are added. The resulting score ranges from 0 to 1, where a value of 1 implies the images are identical.
In real world defogging scenarios, ground truth images are not available.
To address this limitation, no-reference image quality metrics are used to check efficiency of restored images.
\subsection{Fog Aware Density Evaluator (FADE)}

FADE measures the perceptual density of fog by analyzing the statistical distribution of low-level features related to contrast attenuation and visibility loss. Let $\mathbf{f}$ denote a feature vector extracted from the image. The FADE score is computed as
\begin{equation}
\text{FADE} = \sum_{k} w_k \, \phi_k(\mathbf{f}),
\end{equation}
where $\phi_k(\cdot)$ represents fog-related feature responses and $w_k$ are learned weights.
Lower FADE values correspond to clearer images with reduced fog concentration,
making FADE particularly suitable for defogging evaluation without reference images.

\subsection{Contrast Restoration Index (CRI)}

CRI calculates the improvement in contrast after restoration by comparing the dynamic range of intensity values before and after defogging.
It is defined as
\begin{equation}
\text{CRI} =
\frac{I_{\max}^{\text{out}} - I_{\min}^{\text{out}}}
{I_{\max}^{\text{in}} - I_{\min}^{\text{in}}},
\end{equation}
where $I_{\max}^{\text{in}}$ and $I_{\min}^{\text{in}}$ denote the maximum and minimum
intensities of the input foggy image, and
$I_{\max}^{\text{out}}$ and $I_{\min}^{\text{out}}$ correspond to the restored image.
A higher CRI indicates more effective contrast enhancement.

\subsection{Image Entropy}

Entropy calculates the amount of information contained in an image by analyzing the distribution of intensity levels. Let $p_k$ denote the probability of occurrence
of intensity level $k$. The entropy is computed as
\begin{equation}
\text{Entropy} = - \sum_{k} p_k \log_2(p_k).
\end{equation}
Higher value of entropy indicates rich texture details and improved
information content in the restored image.

\subsection{Average Gradient (AG)}

The average gradient quantifies image sharpness by measuring the average magnitude of local intensity variations. It is defined as
\begin{equation}
\text{AG} =
\frac{1}{MN} \sum_{i=1}^{M} \sum_{j=1}^{N}
\sqrt{ \left( \nabla_x I(i,j) \right)^2 + \left( \nabla_y I(i,j) \right)^2 },
\end{equation}
where $\nabla_x I$ and $\nabla_y I$ represent horizontal and vertical gradients,
respectively. A higher value of AG indicates sharper edges and fine details in defogged images.

\begin{figure}[H]
    \centering
    \setlength{\tabcolsep}{1.5pt}
    \renewcommand{\arraystretch}{1.1}

    \begin{tabular}{c c c c c}

    \includegraphics[width=0.18\textwidth]{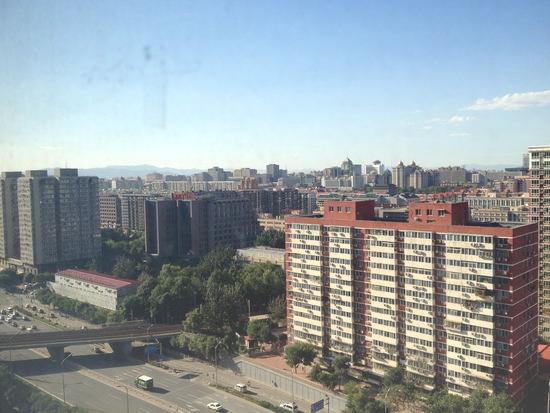} &
    \includegraphics[width=0.18\textwidth]{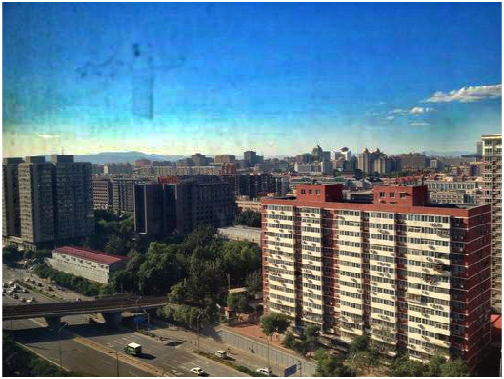} &
    \includegraphics[width=0.18\textwidth]{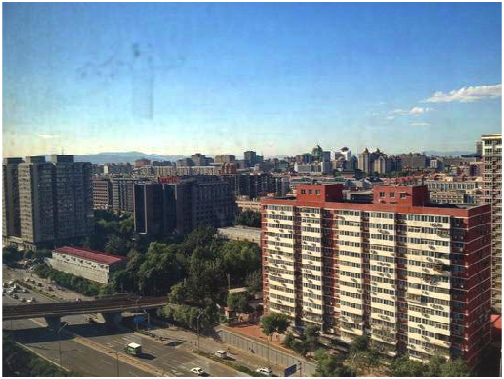} &
    \includegraphics[width=0.18\textwidth]{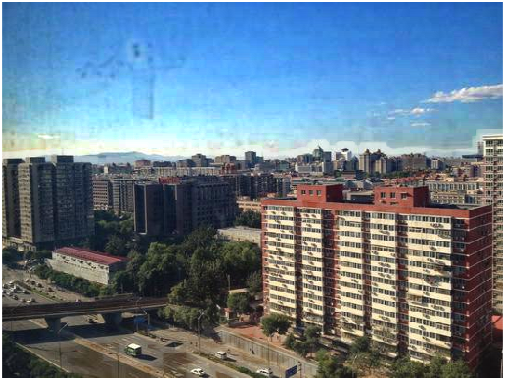} &
    \includegraphics[width=0.18\textwidth]{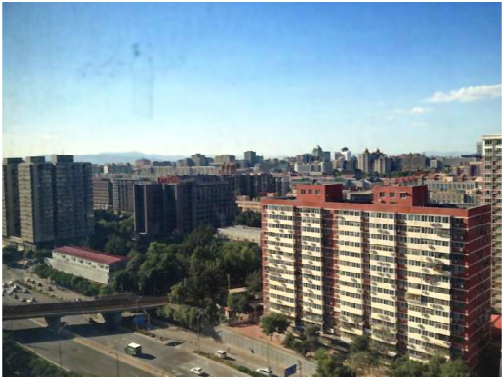} \\[-2pt]

    {\footnotesize Foggy (10\%)} &
    {\footnotesize DCP} &
    {\footnotesize MDCP} &
    {\footnotesize SIDVBM} &
    {\footnotesize Proposed} \\[6pt]

    \includegraphics[width=0.18\textwidth]{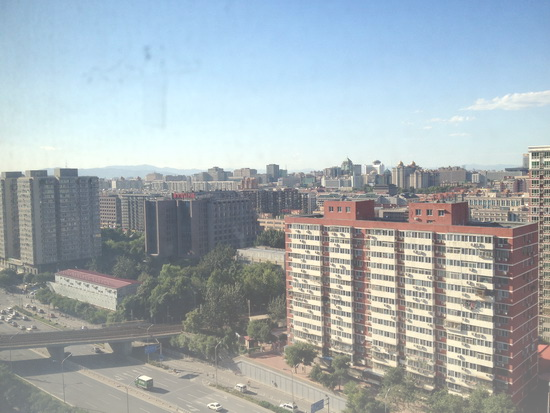} &
    \includegraphics[width=0.18\textwidth]{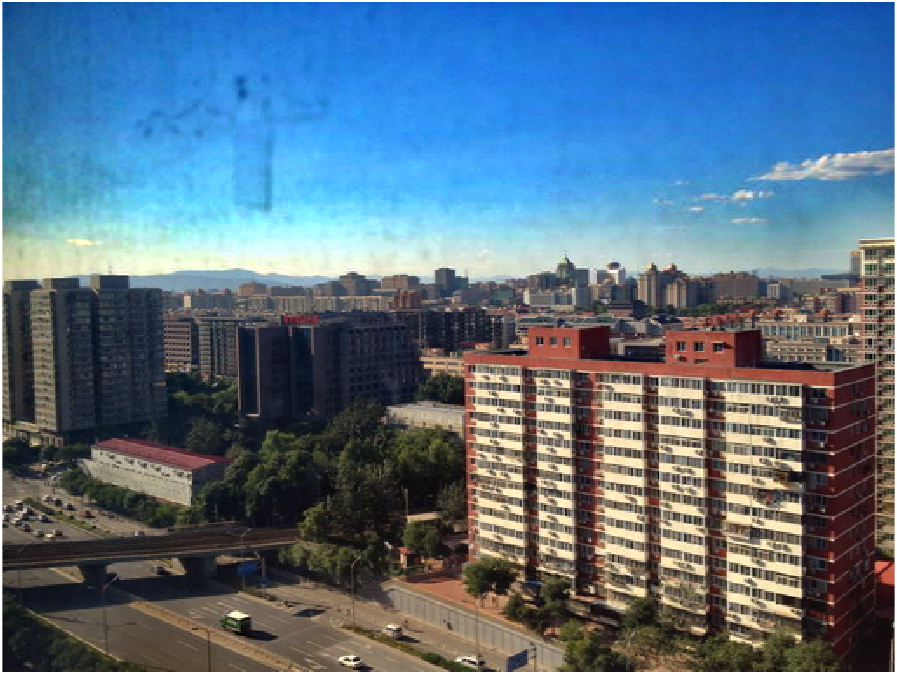} &
    \includegraphics[width=0.18\textwidth]{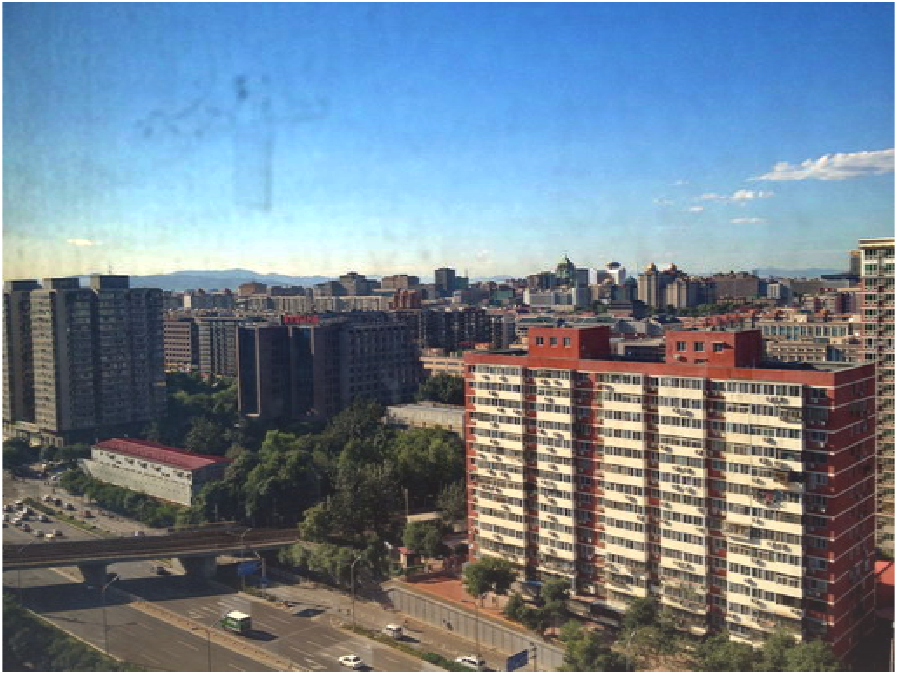} &
    \includegraphics[width=0.18\textwidth]{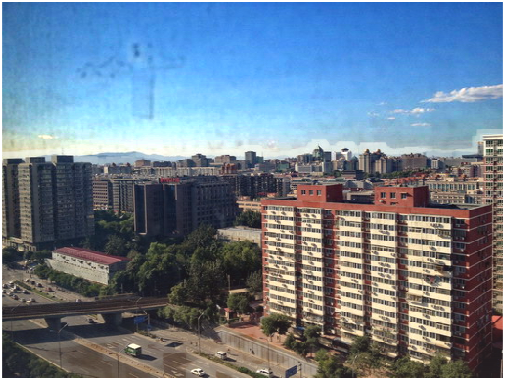} &
    \includegraphics[width=0.18\textwidth]{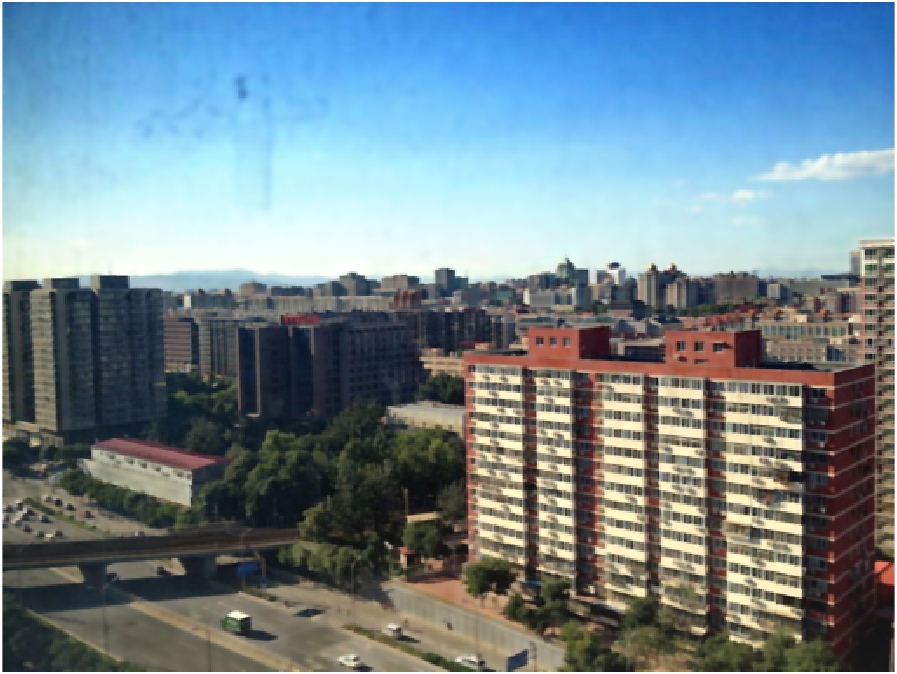} \\[-2pt]

    {\footnotesize Foggy (20\%)} &
    {\footnotesize DCP} &
    {\footnotesize MDCP} &
    {\footnotesize SIDVBM} &
    {\footnotesize Proposed} \\[6pt]

    \includegraphics[width=0.18\textwidth]{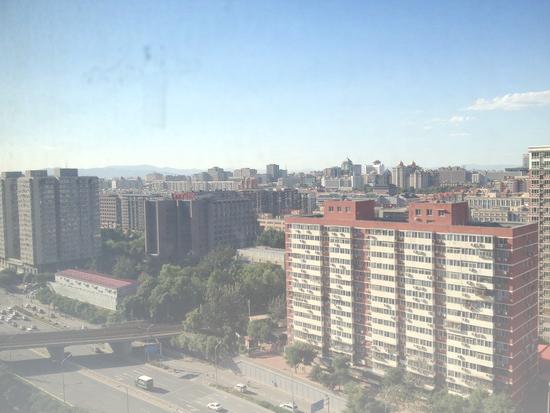} &
    \includegraphics[width=0.18\textwidth]{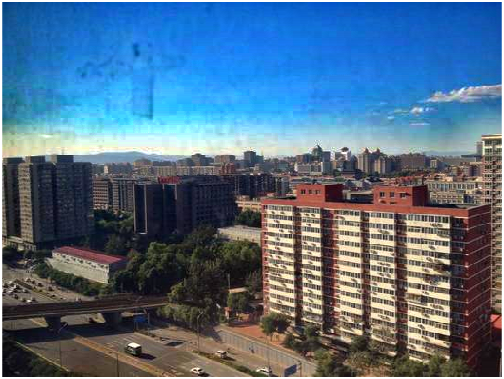} &
    \includegraphics[width=0.18\textwidth]{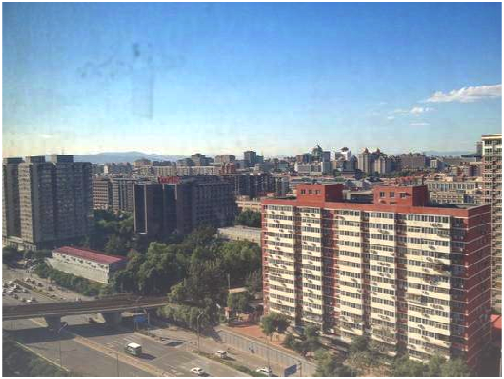} &
    \includegraphics[width=0.18\textwidth]{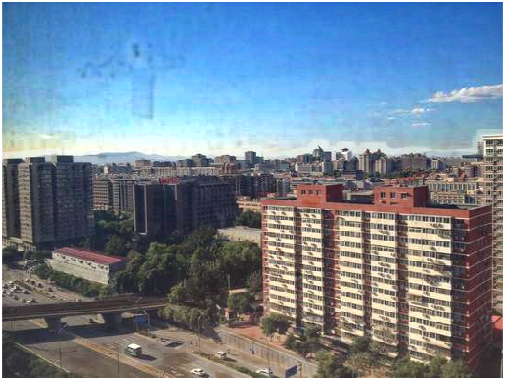} &
    \includegraphics[width=0.18\textwidth]{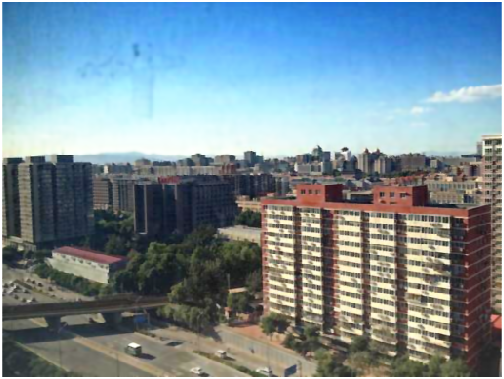} \\[-2pt]

    {\footnotesize Foggy (30\%)} &
    {\footnotesize DCP} &
    {\footnotesize MDCP} &
    {\footnotesize SIDVBM} &
    {\footnotesize Proposed}

    \end{tabular}

    \caption{Visual Comparison of defogging for Img 1. The first column shows the input images, and the second, third, and fourth columns contain the restored versions obtained using DCP\cite{he2011single}, MDCP\cite{salazar2019fast}, SIDVBM\cite{liu2019unified}, and the proposed Model.}

    \label{fig:defogging_comparison_img1}
\end{figure}
\begin{figure}[H]
    \centering
    \setlength{\tabcolsep}{1.5pt}
    \renewcommand{\arraystretch}{1.1}

    \begin{tabular}{c c c c c}

    \includegraphics[width=0.18\textwidth]{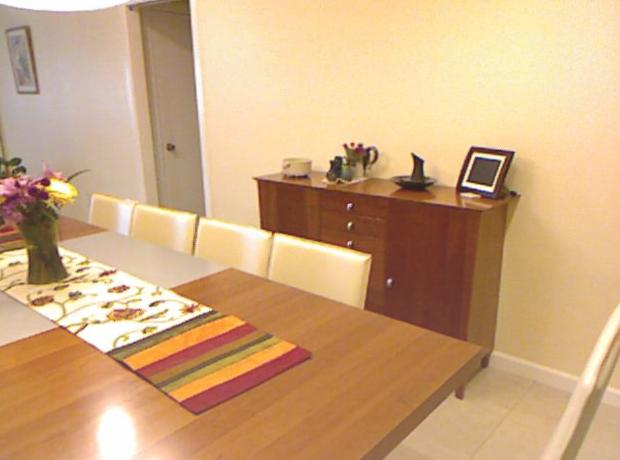} &
    \includegraphics[width=0.18\textwidth]{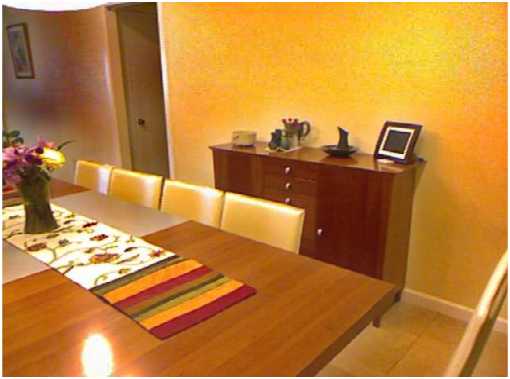} &
    \includegraphics[width=0.18\textwidth]{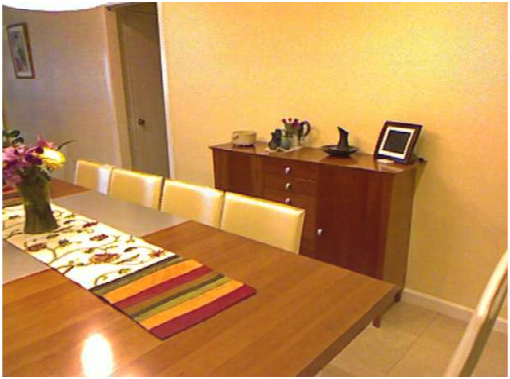} &
    \includegraphics[width=0.18\textwidth]{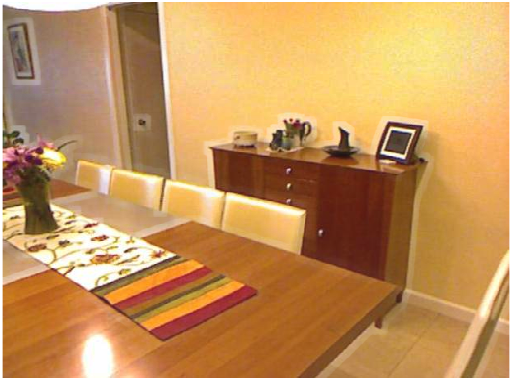} &
    \includegraphics[width=0.18\textwidth]{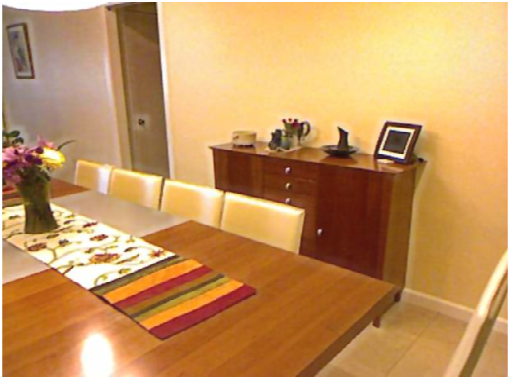} \\[-2pt]

    {\footnotesize Foggy (10\%)} &
    {\footnotesize DCP} &
    {\footnotesize MDCP} &
    {\footnotesize SIDVBM} &
    {\footnotesize Proposed} \\[6pt]

    \includegraphics[width=0.18\textwidth]{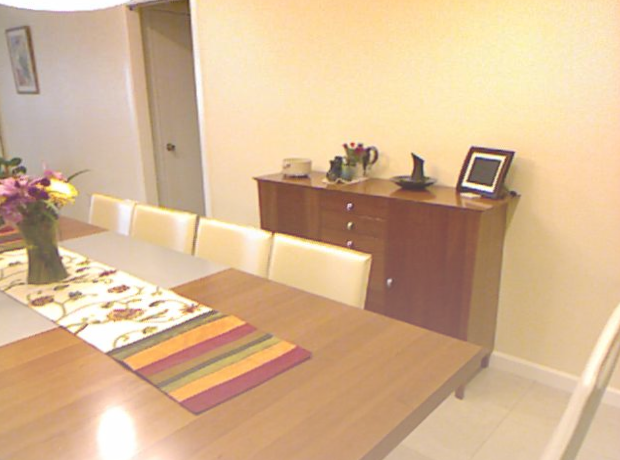} &
    \includegraphics[width=0.18\textwidth]{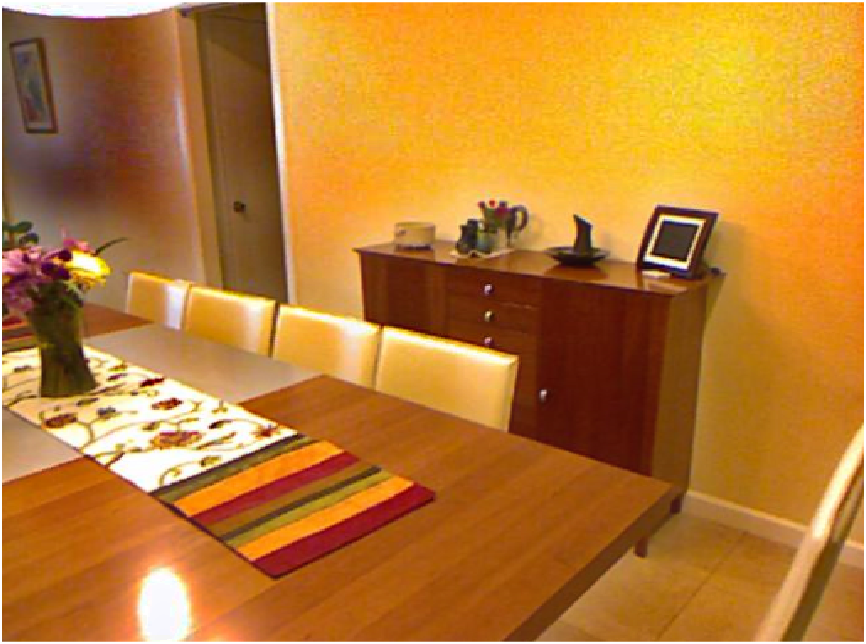} &
    \includegraphics[width=0.18\textwidth]{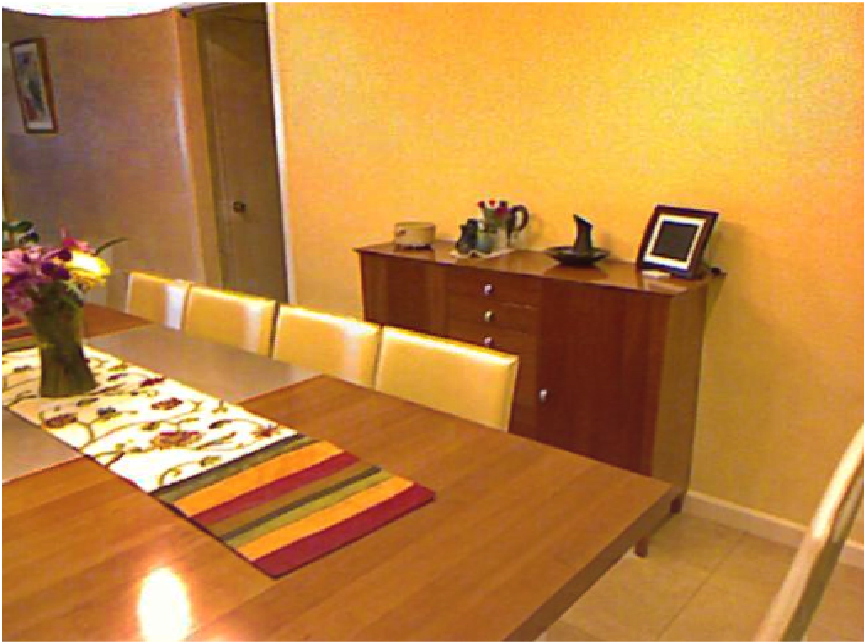} &
    \includegraphics[width=0.18\textwidth]{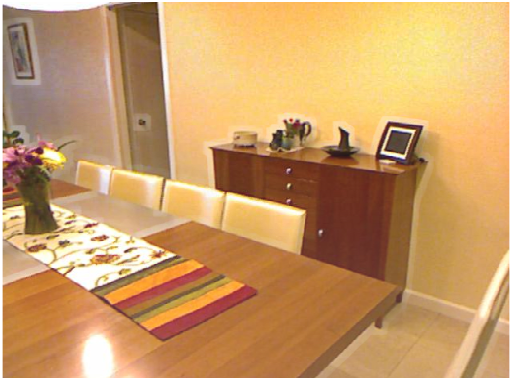} &
    \includegraphics[width=0.18\textwidth]{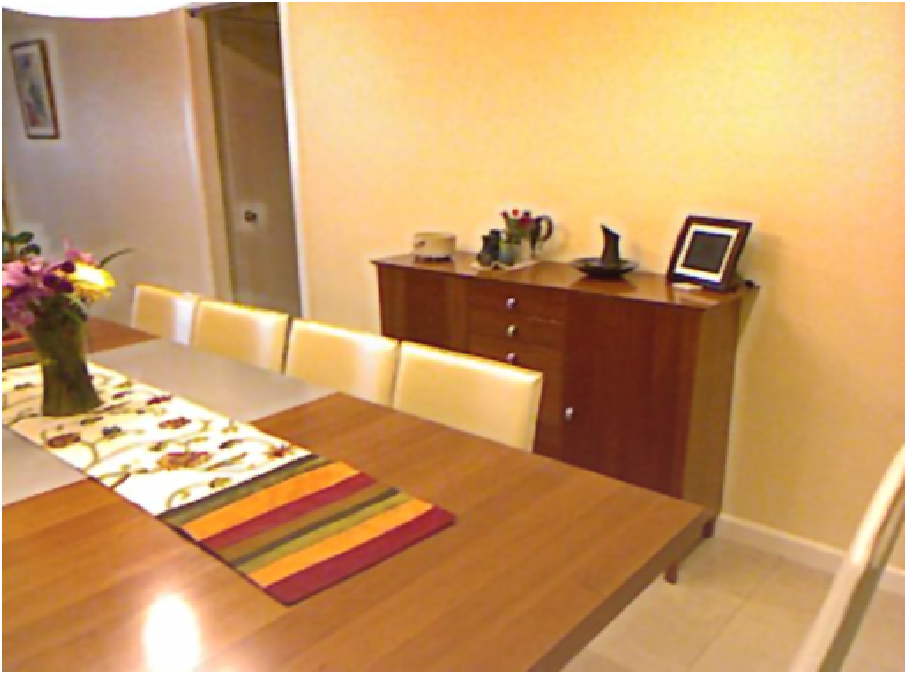} \\[-2pt]

    {\footnotesize Foggy (20\%)} &
    {\footnotesize DCP} &
    {\footnotesize MDCP} &
    {\footnotesize SIDVBM} &
    {\footnotesize Proposed} \\[6pt]

    \includegraphics[width=0.18\textwidth]{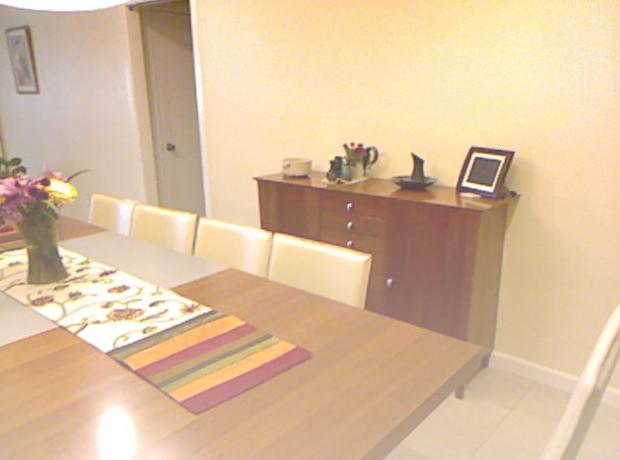} &
    \includegraphics[width=0.18\textwidth]{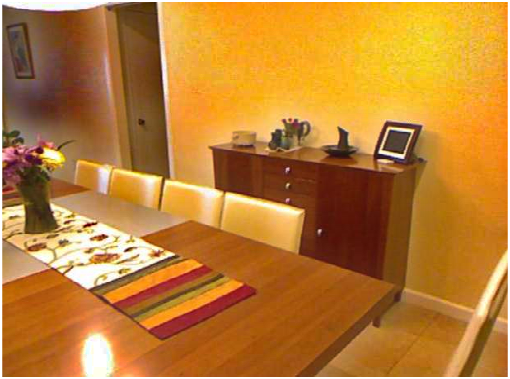} &
    \includegraphics[width=0.18\textwidth]{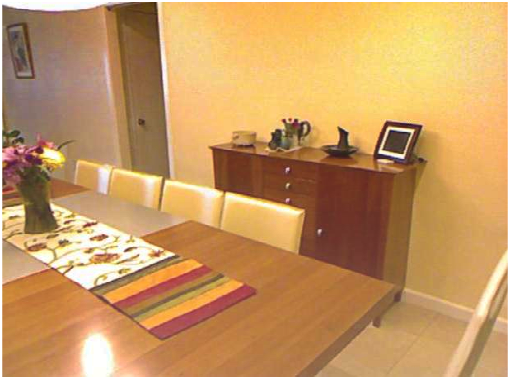} &
    \includegraphics[width=0.18\textwidth]{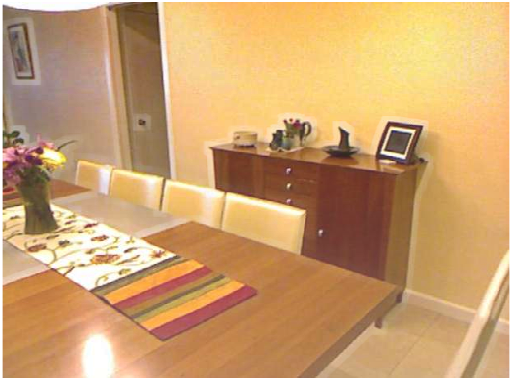} &
    \includegraphics[width=0.18\textwidth]{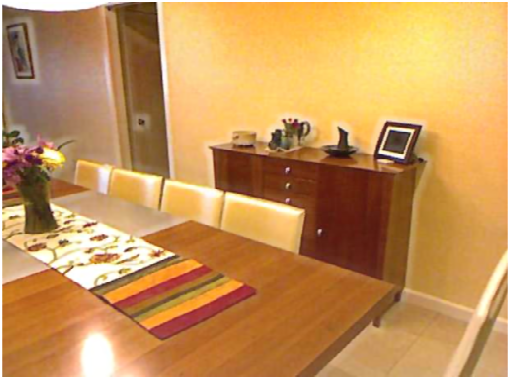} \\[-2pt]

    {\footnotesize Foggy (30\%)} &
    {\footnotesize DCP} &
    {\footnotesize MDCP} &
    {\footnotesize SIDVBM} &
    {\footnotesize Proposed}

    \end{tabular}

    \caption{Visual Comparison of defogging for Img 6. The first column shows the input images, and the second, third, and fourth columns contain the restored versions obtained using DCP\cite{he2011single}, MDCP\cite{salazar2019fast}, SIDVBM\cite{liu2019unified}, and the proposed model.}

    \label{fig:defogging_comparison_img6}
\end{figure}

Table~\ref{tab:defogging_mse_ssim} presents a detailed quantitative evaluation of the defogging performance using Mean Squared Error (MSE) and Structural Similarity Index (SSIM) for six color images under different fog densities. In MSE smaller values indicate higher defogging image reconstruction accuracy, while SSIM evaluates the preservation of structural and perceptual information. The higher values of SSIM indicating better visual similarity. From the tablee~\ref{tab:defogging_mse_ssim}, it is clearly observed that the proposed fourth-order PDE-based defogging method outperforms the DCP, MDCP, and SIDVBM approaches by achieving lower MSE and higher SSIM values for all test images and fog levels. This improvement becomes significant enhancement under moderate and heavy fog conditions (20\% and 30\%), where conventional methods fails to captures fine details and structure. Also the traditional method enhance the colors so we also see the color distortion, while our proposed model give balace betwwen colors restoration and defogging. These quantitative findings are well in visual comparisons shown in Figures~\ref{fig:defogging_comparison_img1}~\ref{fig:defogging_comparison_img6}, where the proposed approach produces clearer images with balanced color contrast, sharper edges, and more natural color appearance compared to the DCP, MDCP, and SIDVBM approaches. To check the defogging performance in situations where ground truth images are not available,  no-reference image quality metrics are used. The quantitative results are summarized in Table~\ref{tab:defogging_qualitative_metrics} using FADE, Color Restoration Index (CRI), Entropy, and Average Gradient (AG) for five representative non reference color images. In FADE metric lower values indicate better fog removal, while CRI evaluates the quality of color restoration. Entropy gives us the richness of image information, and AG measures edge strength and sharpness. Results demonstrate that the original foggy images give very high FADE values along with low AG values, confirming the presence of dense fog and poor edge visibility. After applying defogging algorithms, all compared methods show a noticeable reduction in FADE values, indicating fog suppression. However, our proposed fourth-order PDE-based method gives the lowest FADE values across all non reference  images, demonstrating its superior capability in removing fog more than traditional DCP, MDCP, and SIDVBM methods.
Also, the proposed approach gives higher CRI values for all non reference images, which indicates more accurate color recovery. In case of entropy values the proposed method are also better performing than DCP and MDCP methodsin cases of non reference images. Furthermore, the increased AG values indicate enhanced edge sharpness, which are crucial for improving overall visibility. These quantitative metrics values improvements shows that the proposed method not only reduces fog effectively but also preserves important image details and enhances perceptual quality of images.

\begin{figure}[H]
    \centering
    \setlength{\tabcolsep}{1.5pt}
    \renewcommand{\arraystretch}{1.1}

    \begin{tabular}{c c c c c}

    \includegraphics[width=0.18\textwidth]{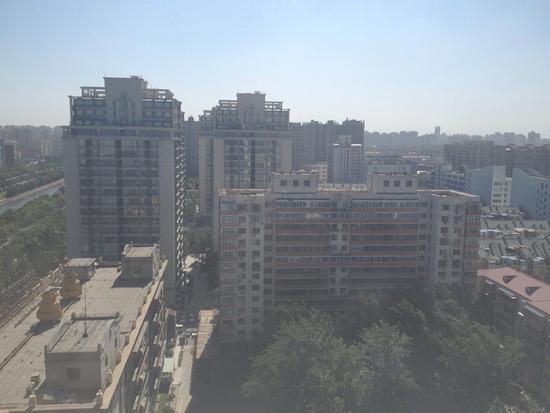} &
    \includegraphics[width=0.18\textwidth]{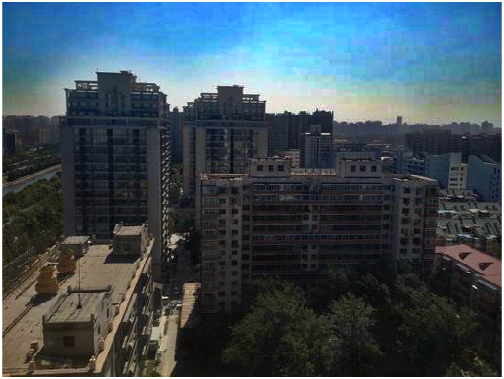} &
    \includegraphics[width=0.18\textwidth]{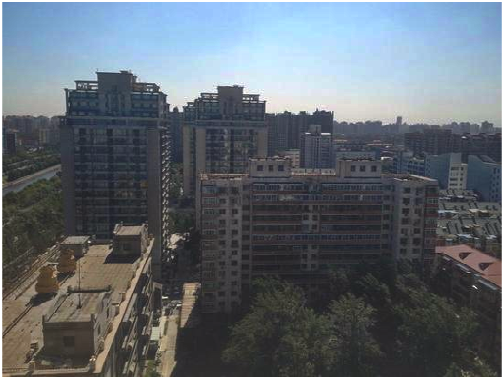} &
    \includegraphics[width=0.18\textwidth]{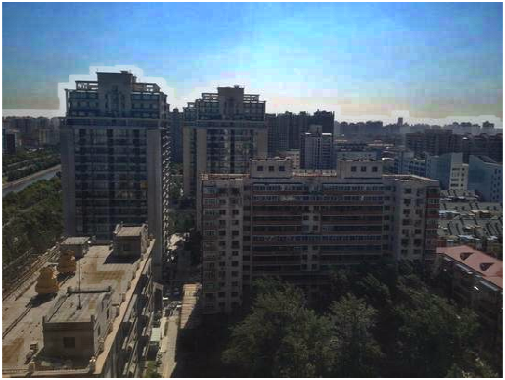} &
    \includegraphics[width=0.18\textwidth]{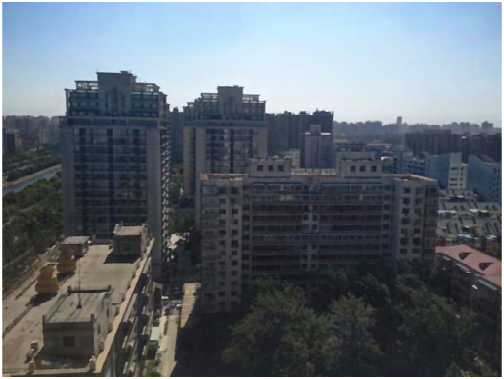} \\[-2pt]

    {\footnotesize Foggy (10\%)} &
    {\footnotesize DCP} &
    {\footnotesize MDCP} &
    {\footnotesize SIDVBM} &
    {\footnotesize Proposed} \\[6pt]

    \includegraphics[width=0.18\textwidth]{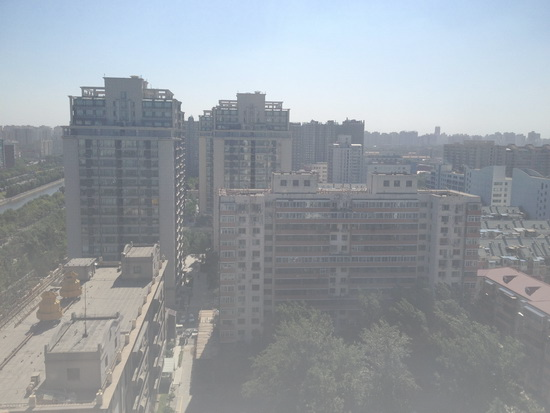} &
    \includegraphics[width=0.18\textwidth]{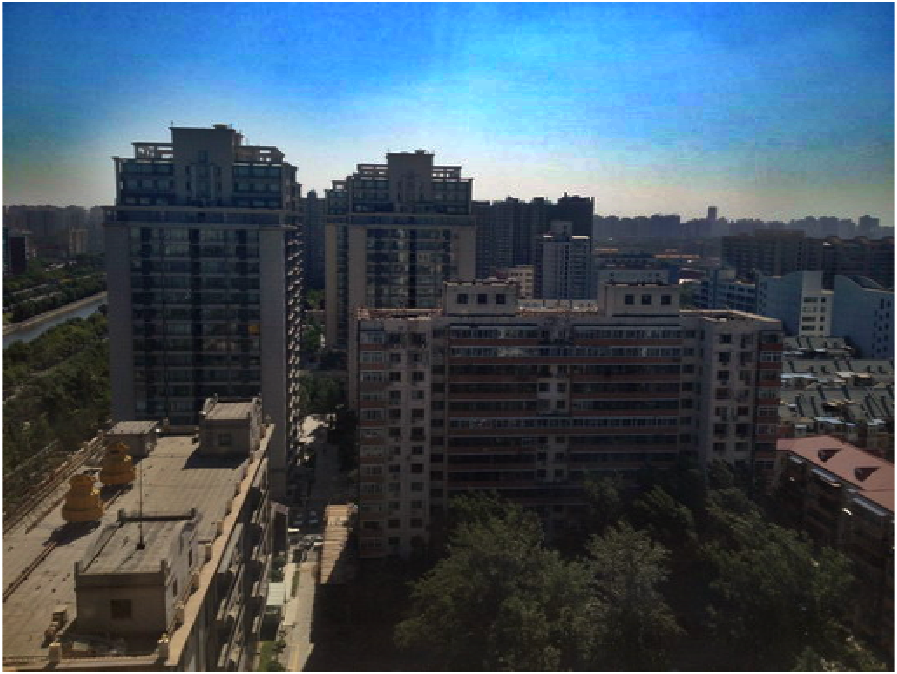} &
    \includegraphics[width=0.18\textwidth]{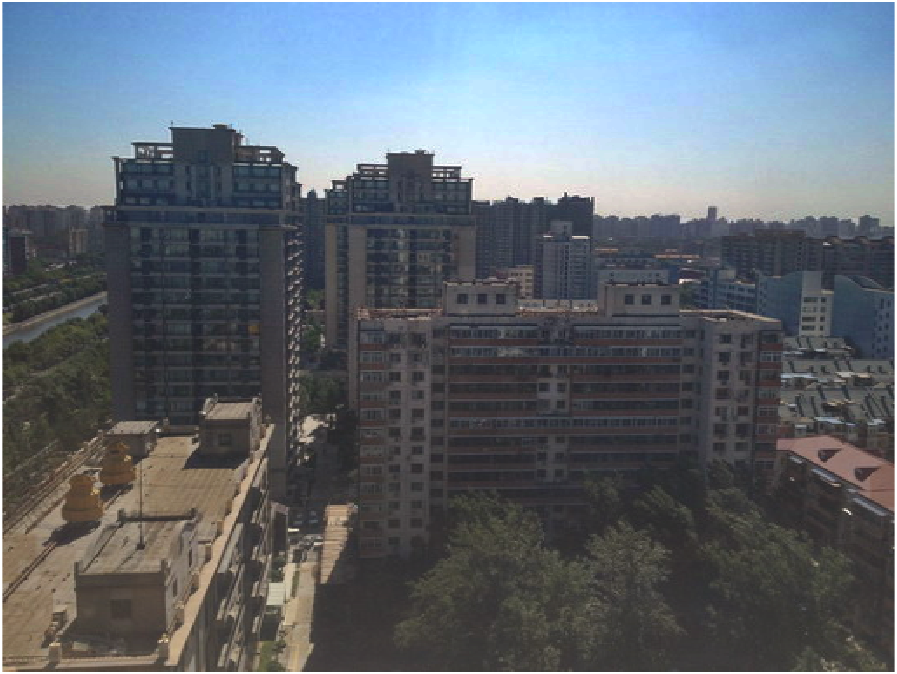} &
    \includegraphics[width=0.18\textwidth]{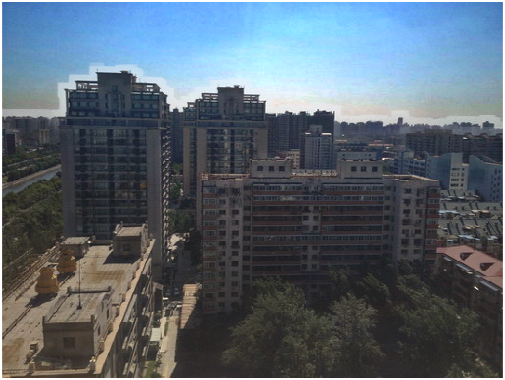} &
    \includegraphics[width=0.18\textwidth]{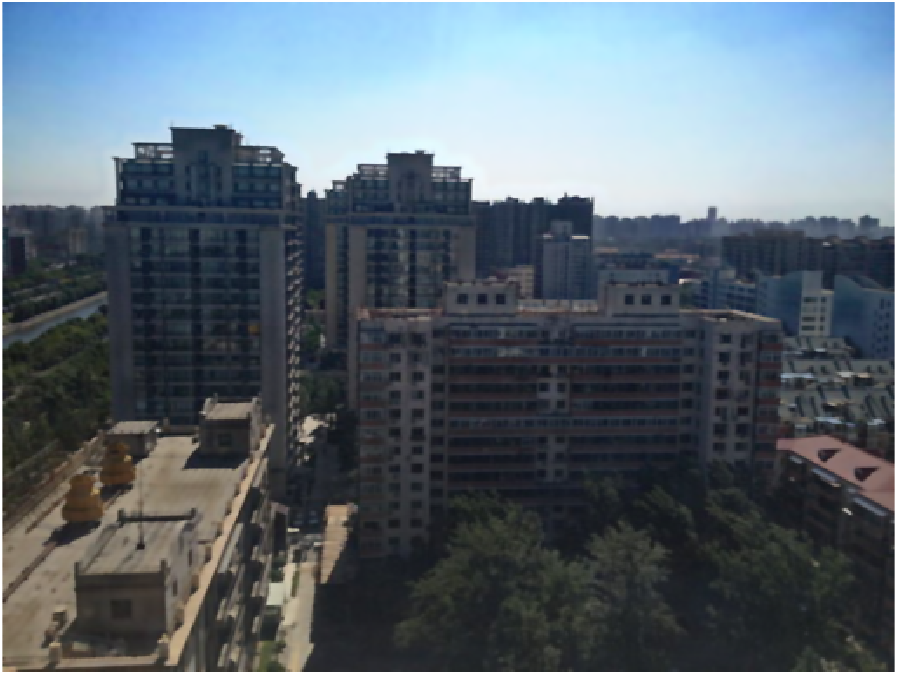} \\[-2pt]

    {\footnotesize Foggy (20\%)} &
    {\footnotesize DCP} &
    {\footnotesize MDCP} &
    {\footnotesize SIDVBM} &
    {\footnotesize Proposed} \\[6pt]

    \includegraphics[width=0.18\textwidth]{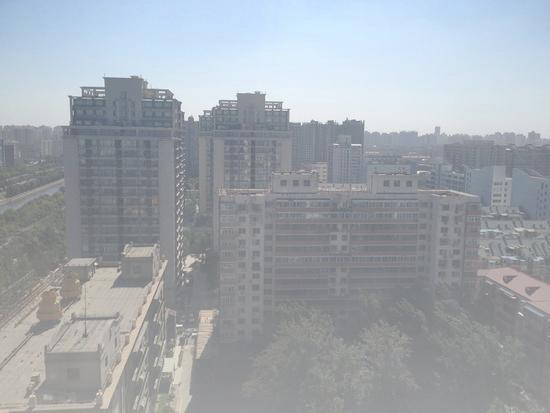} &
    \includegraphics[width=0.18\textwidth]{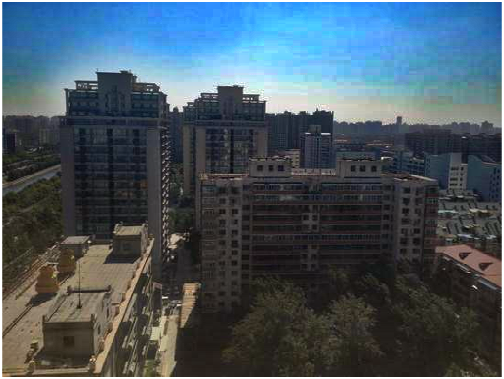} &
    \includegraphics[width=0.18\textwidth]{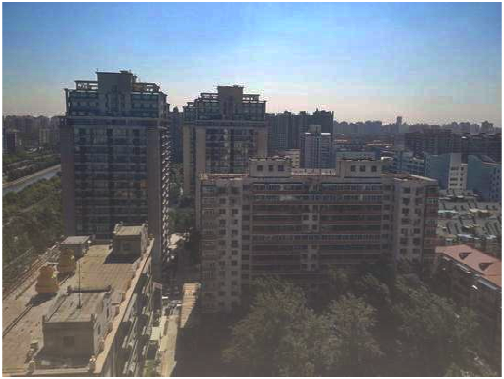} &
    \includegraphics[width=0.18\textwidth]{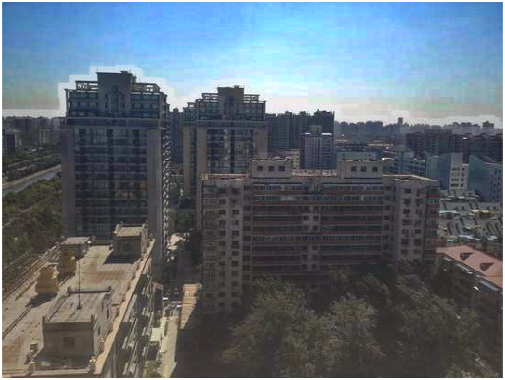} &
    \includegraphics[width=0.18\textwidth]{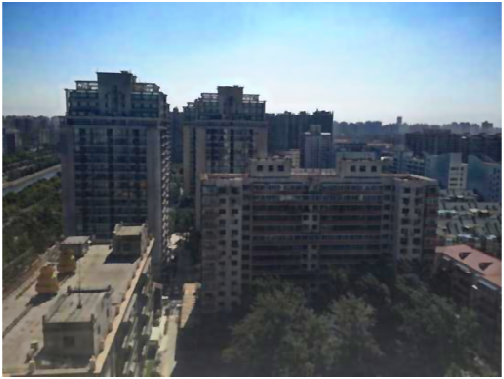} \\[-2pt]

    {\footnotesize Foggy (30\%)} &
    {\footnotesize DCP} &
    {\footnotesize MDCP} &
    {\footnotesize SIDVBM} &
    {\footnotesize Proposed}

    \end{tabular}

    \caption{Visual Comparison of defogging for Img 2. The first column shows the input images, and the second, third, and fourth columns contain the restored versions obtained using DCP\cite{he2011single}, MDCP\cite{salazar2019fast}, SIDVBM\cite{liu2019unified}, and the proposed Model.}

    \label{fig:defogging_comparison_img2}
\end{figure}

\begin{figure}[H]
    \centering
    \setlength{\tabcolsep}{1.5pt}
    \renewcommand{\arraystretch}{1.1}

    \begin{tabular}{c c c c c}

    \includegraphics[width=0.18\textwidth]{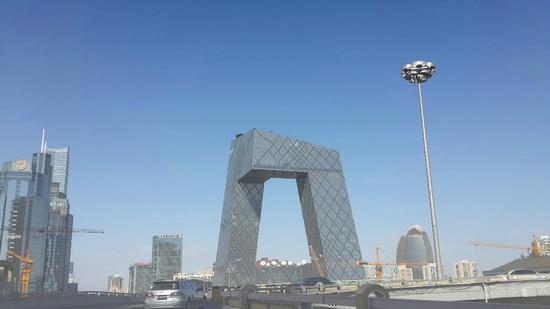} &
    \includegraphics[width=0.18\textwidth]{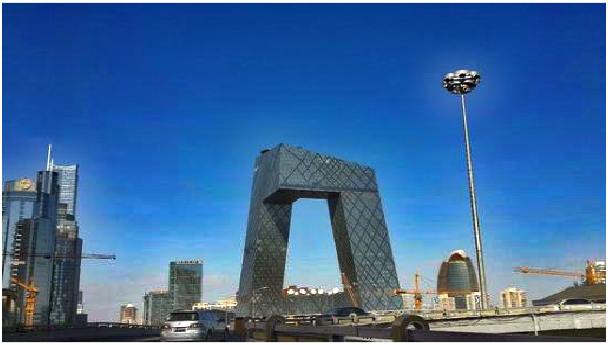} &
    \includegraphics[width=0.18\textwidth]{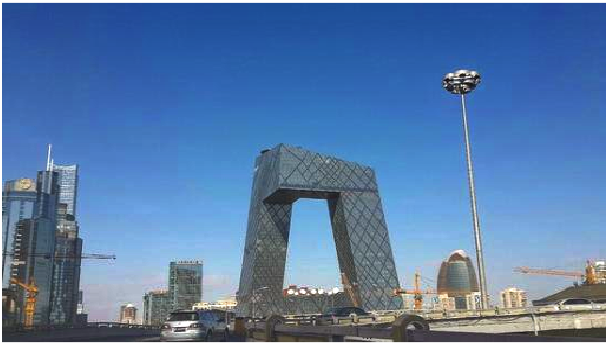} &
    \includegraphics[width=0.18\textwidth]{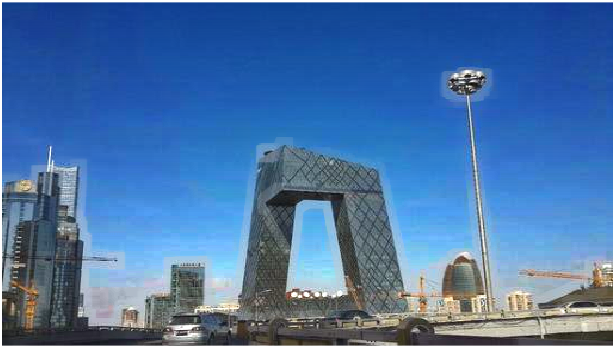} &
    \includegraphics[width=0.18\textwidth]{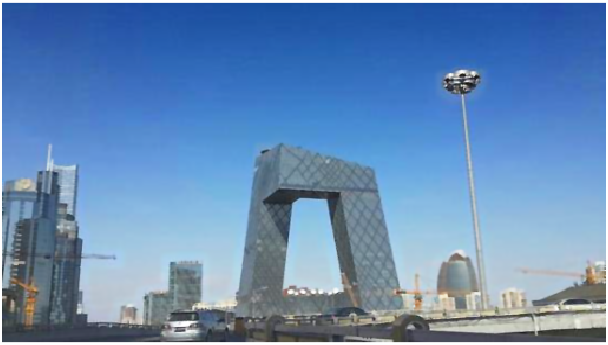} \\[-2pt]

    {\footnotesize Foggy (10\%)} &
    {\footnotesize DCP} &
    {\footnotesize MDCP} &
    {\footnotesize SIDVBM} &
    {\footnotesize Proposed} \\[6pt]

    \includegraphics[width=0.18\textwidth]{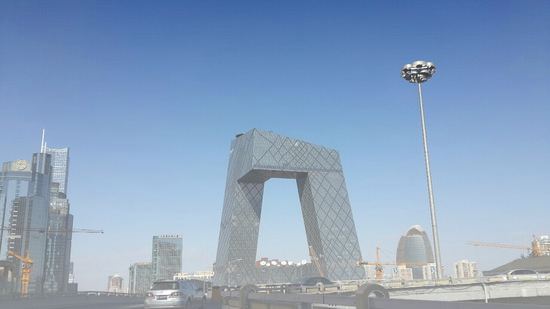} &
    \includegraphics[width=0.18\textwidth]{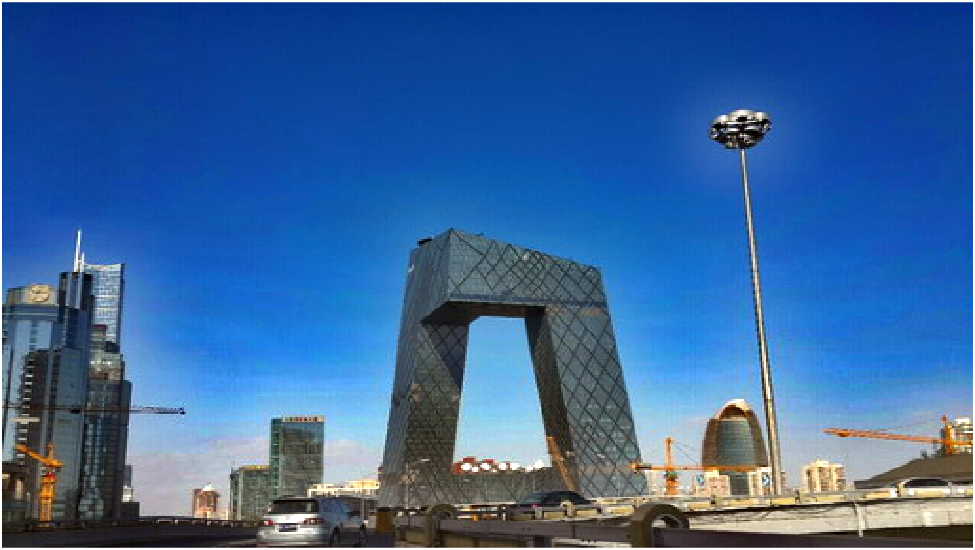} &
    \includegraphics[width=0.18\textwidth]{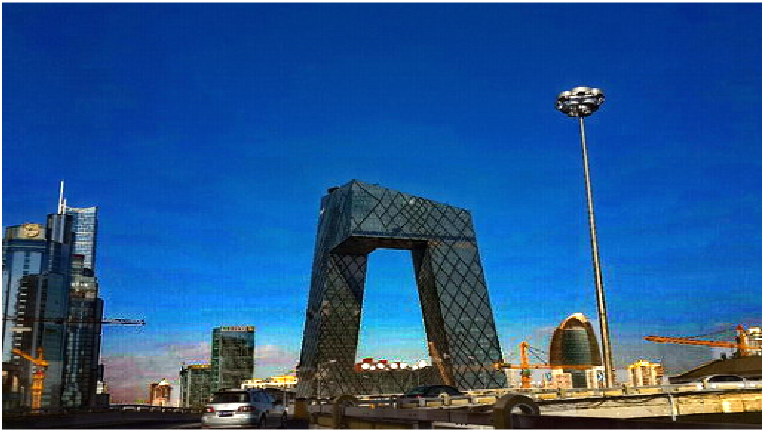} &
    \includegraphics[width=0.18\textwidth]{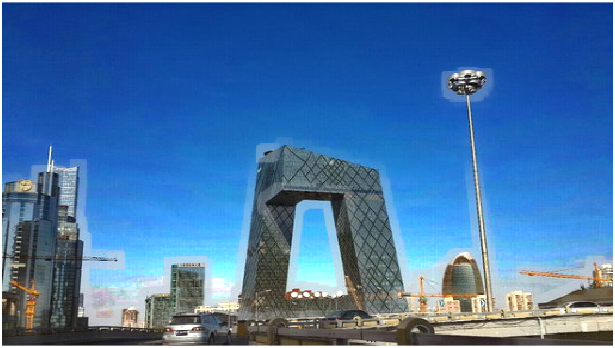} &
    \includegraphics[width=0.18\textwidth]{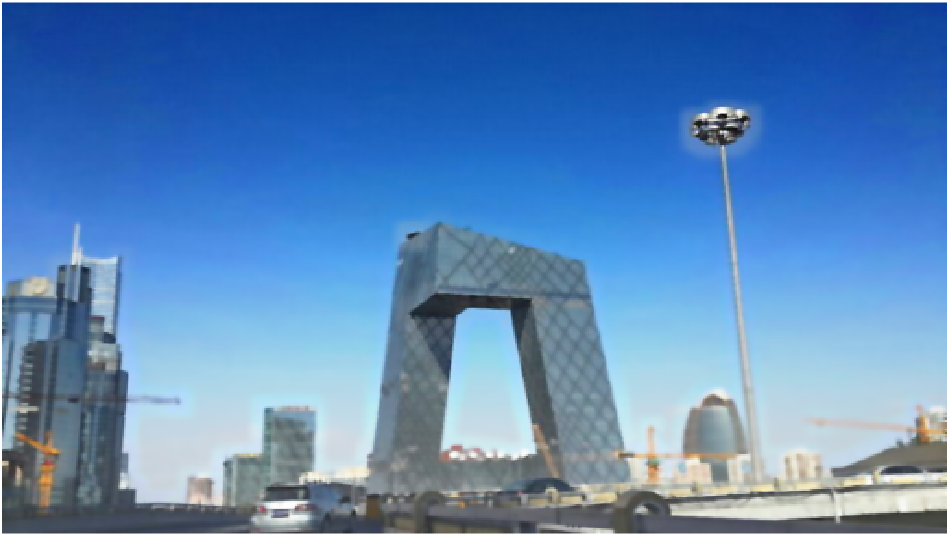} \\[-2pt]

    {\footnotesize Foggy (20\%)} &
    {\footnotesize DCP} &
    {\footnotesize MDCP} &
    {\footnotesize SIDVBM} &
    {\footnotesize Proposed} \\[6pt]

    \includegraphics[width=0.18\textwidth]{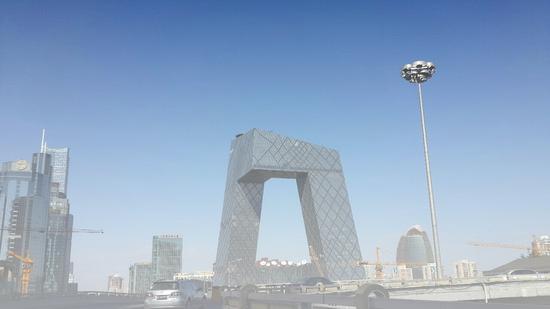} &
    \includegraphics[width=0.18\textwidth]{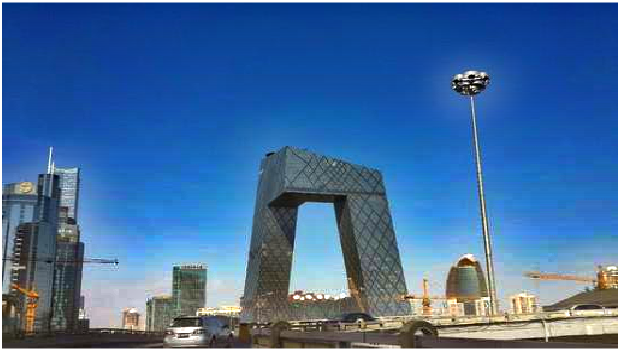} &
    \includegraphics[width=0.18\textwidth]{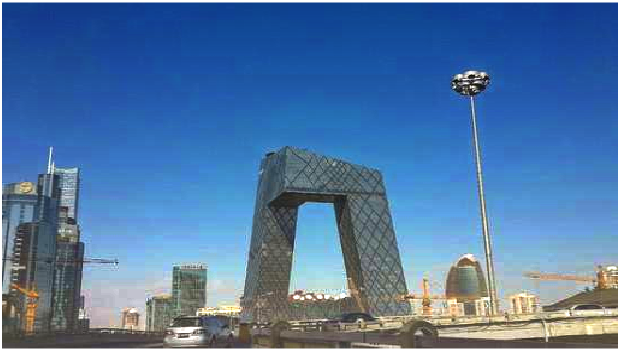} &
    \includegraphics[width=0.18\textwidth]{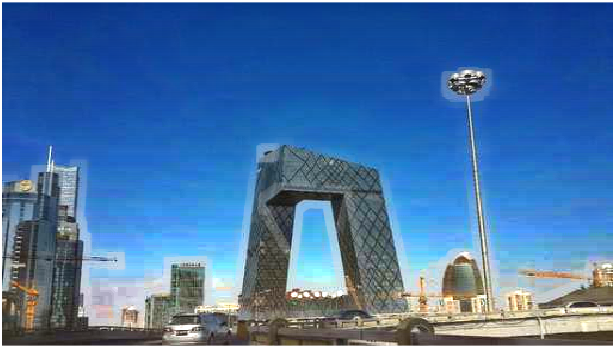} &
    \includegraphics[width=0.18\textwidth]{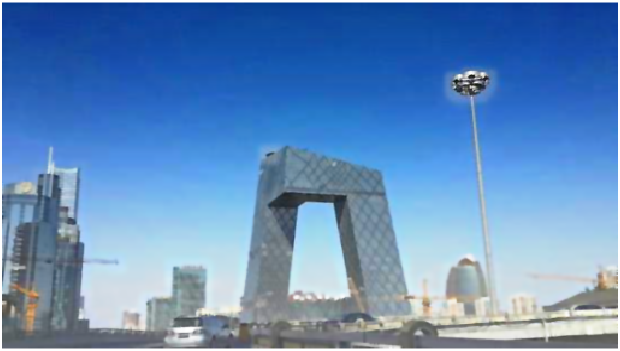} \\[-2pt]

    {\footnotesize Foggy (30\%)} &
    {\footnotesize DCP} &
    {\footnotesize MDCP} &
    {\footnotesize SIDVBM} &
    {\footnotesize Proposed}

    \end{tabular}

    \caption{Visual Comparison of defogging for Img 3. The first column shows the input images, and the second, third, and fourth columns contain the restored versions obtained using DCP\cite{he2011single}, MDCP\cite{salazar2019fast}, SIDVBM\cite{liu2019unified}, and the proposed Model.}

    \label{fig:defogging_comparison_img3}
\end{figure}
\begin{figure}[H]
    \centering
    \setlength{\tabcolsep}{1.5pt}
    \renewcommand{\arraystretch}{1.1}
     \begin{tabular}{c c c c c}

    \includegraphics[width=0.18\textwidth]{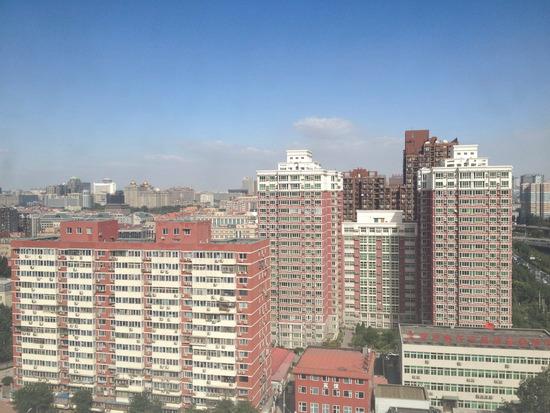} &
    \includegraphics[width=0.18\textwidth]{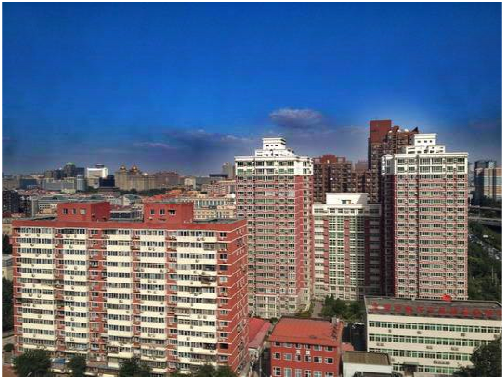} &
    \includegraphics[width=0.18\textwidth]{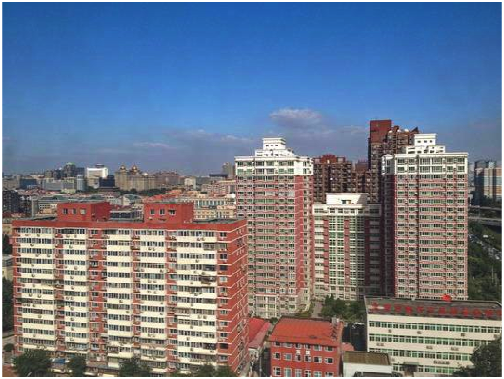} &
    \includegraphics[width=0.18\textwidth]{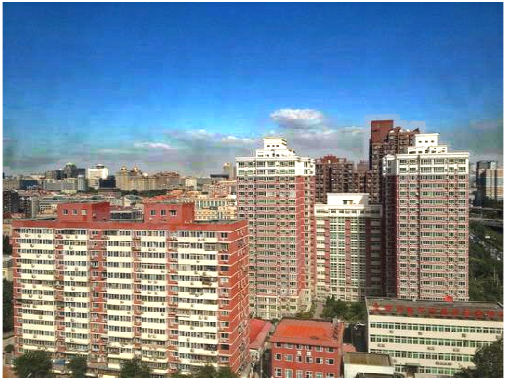} &
    \includegraphics[width=0.18\textwidth]{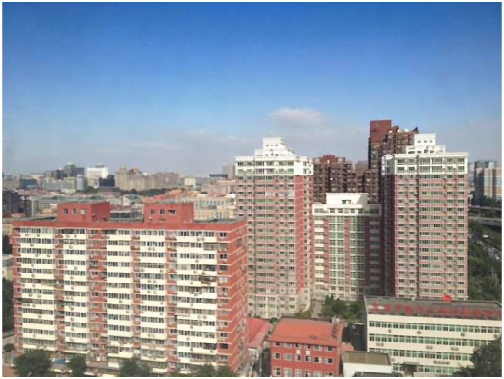} \\[-2pt]

    {\footnotesize Foggy (10\%)} &
    {\footnotesize DCP} &
    {\footnotesize MDCP} &
    {\footnotesize SIDVBM} &
    {\footnotesize Proposed} \\[6pt]

    \includegraphics[width=0.18\textwidth]{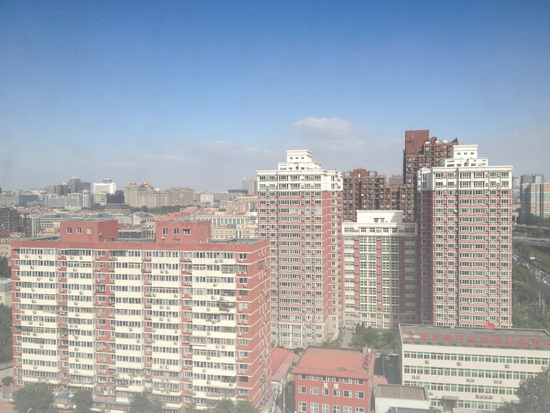} &
    \includegraphics[width=0.18\textwidth]{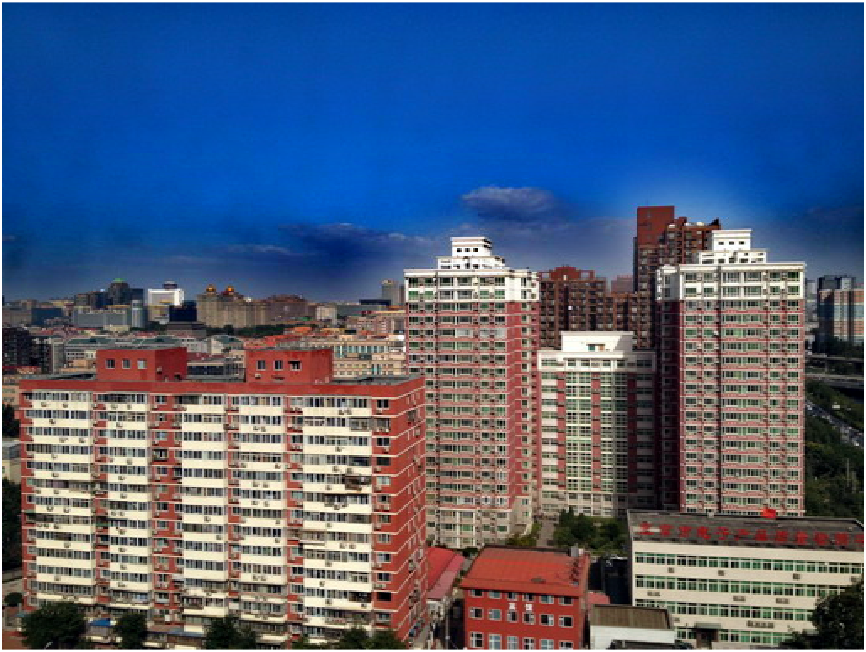} &
    \includegraphics[width=0.18\textwidth]{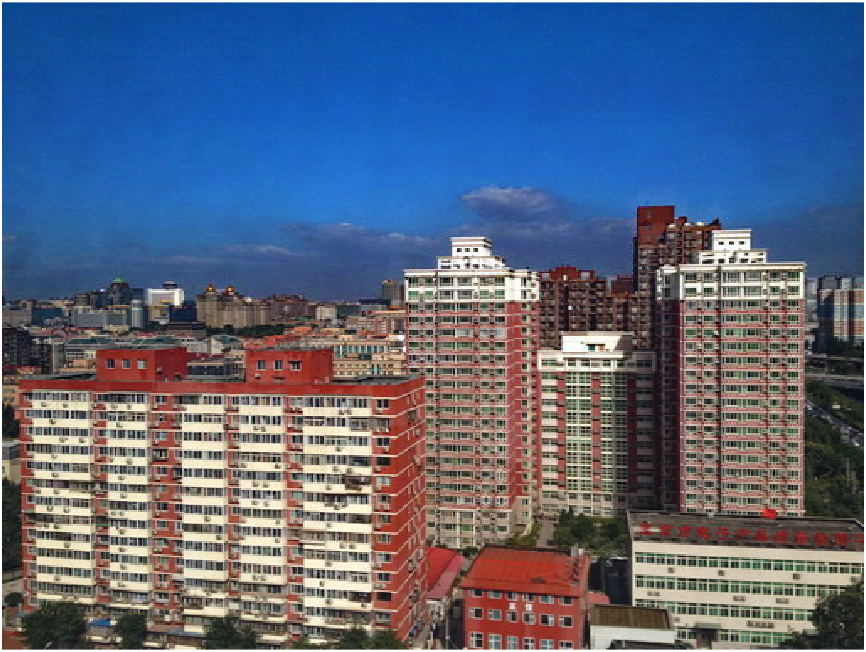} &
    \includegraphics[width=0.18\textwidth]{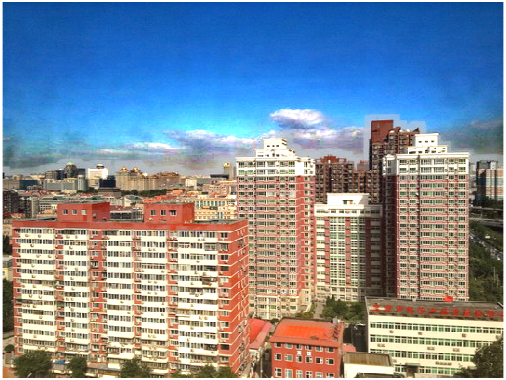} &
    \includegraphics[width=0.18\textwidth]{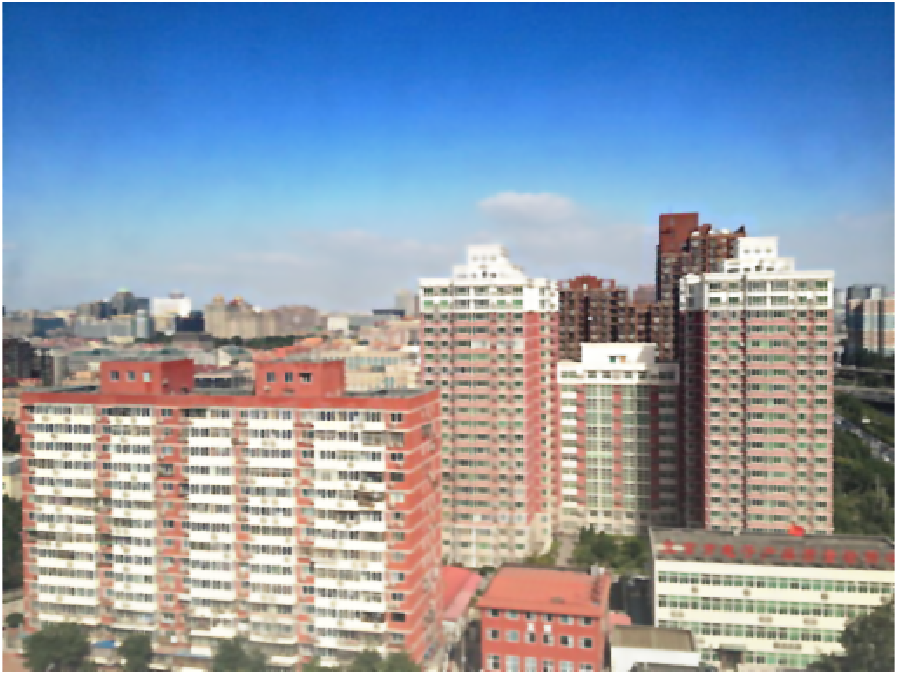} \\[-2pt]

    {\footnotesize Foggy (20\%)} &
    {\footnotesize DCP} &
    {\footnotesize MDCP} &
    {\footnotesize SIDVBM} &
    {\footnotesize Proposed} \\[6pt]

    \includegraphics[width=0.18\textwidth]{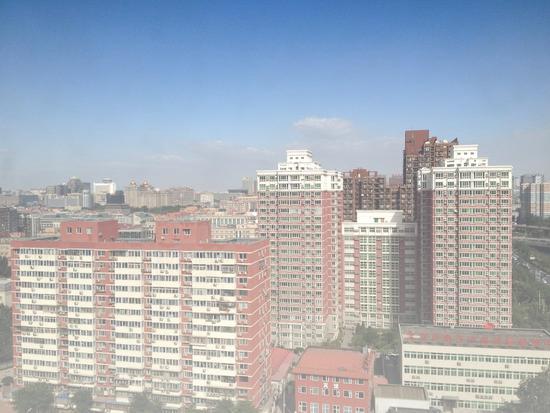} &
    \includegraphics[width=0.18\textwidth]{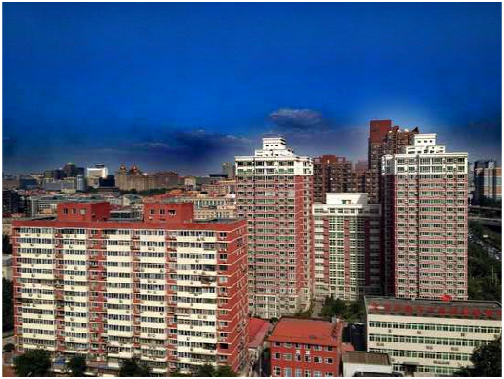} &
    \includegraphics[width=0.18\textwidth]{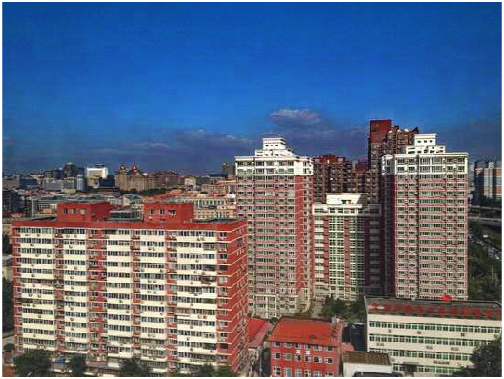} &
    \includegraphics[width=0.18\textwidth]{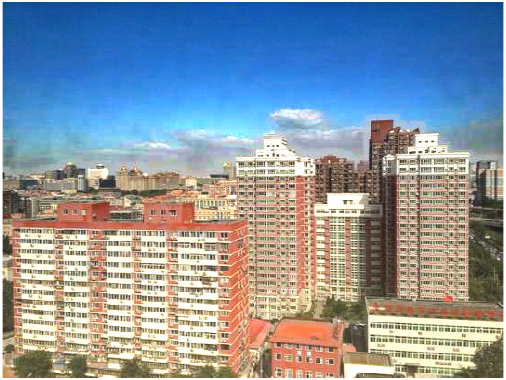} &
    \includegraphics[width=0.18\textwidth]{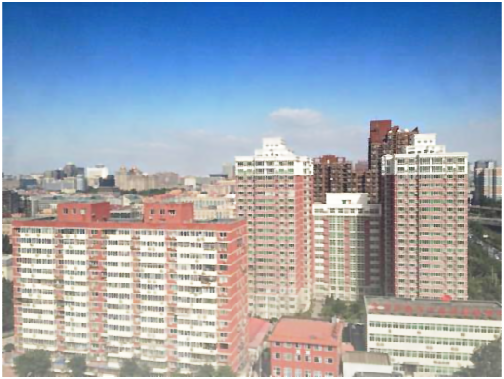} \\[-2pt]

    {\footnotesize Foggy (30\%)} &
    {\footnotesize DCP} &
    {\footnotesize MDCP} &
    {\footnotesize SIDVBM} &
    {\footnotesize Proposed}

    \end{tabular}

    \caption{Visual Comparison of defogging for Img 4. The first column shows the input images, and the second, third, and fourth columns contain the restored versions obtained using DCP\cite{he2011single}, MDCP\cite{salazar2019fast}, SIDVBM\cite{liu2019unified}, and the proposed Model.}

    \label{fig:defogging_comparison_img4}
\end{figure}

\begin{figure}[H]
    \centering
    \setlength{\tabcolsep}{1.5pt}
    \renewcommand{\arraystretch}{1.1}

    \begin{tabular}{c c c c c}

    \includegraphics[width=0.18\textwidth]{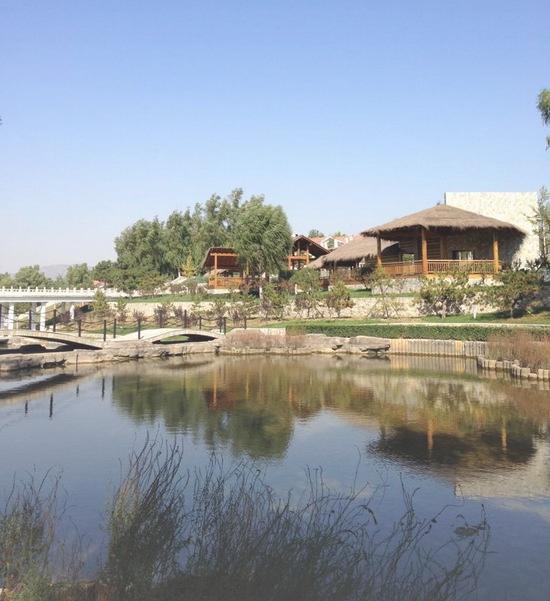} &
    \includegraphics[width=0.18\textwidth]{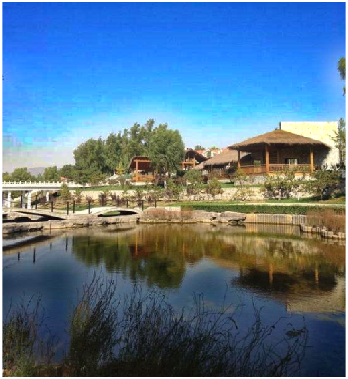} &
    \includegraphics[width=0.18\textwidth]{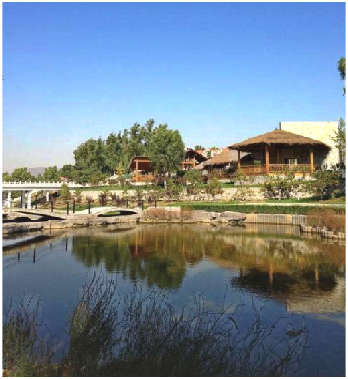} &
    \includegraphics[width=0.18\textwidth]{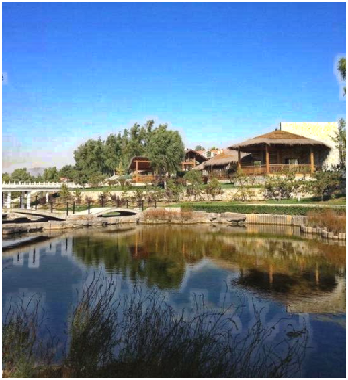} &
    \includegraphics[width=0.18\textwidth]{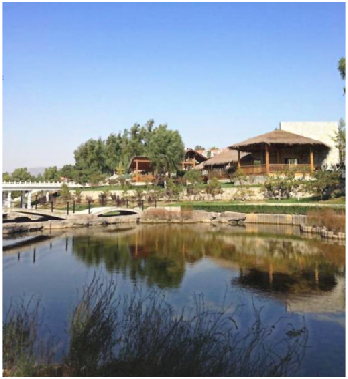} \\[-2pt]

    {\footnotesize Foggy (10\%)} &
    {\footnotesize DCP} &
    {\footnotesize MDCP} &
    {\footnotesize SIDVBM} &
    {\footnotesize Proposed} \\[6pt]

    \includegraphics[width=0.18\textwidth]{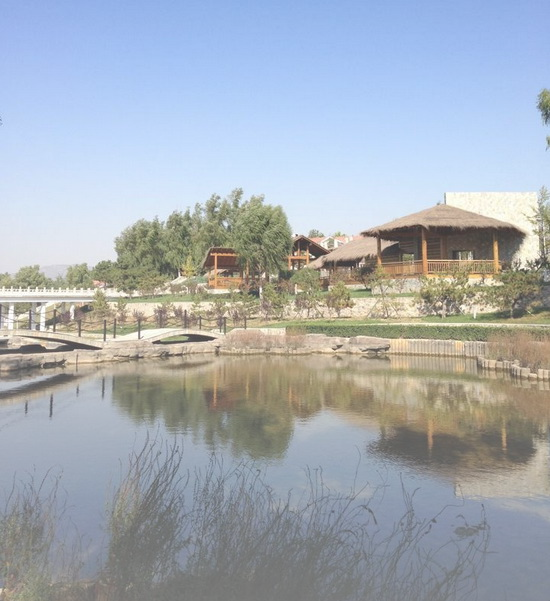} &
    \includegraphics[width=0.18\textwidth]{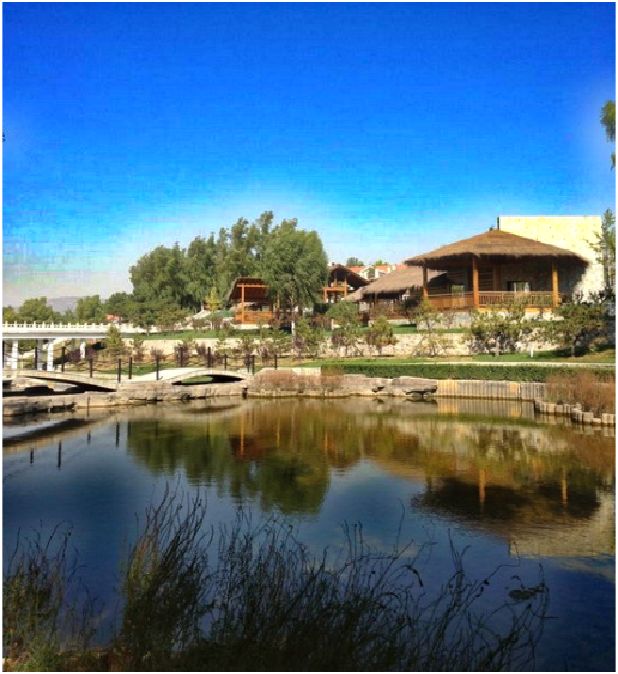} &
    \includegraphics[width=0.18\textwidth]{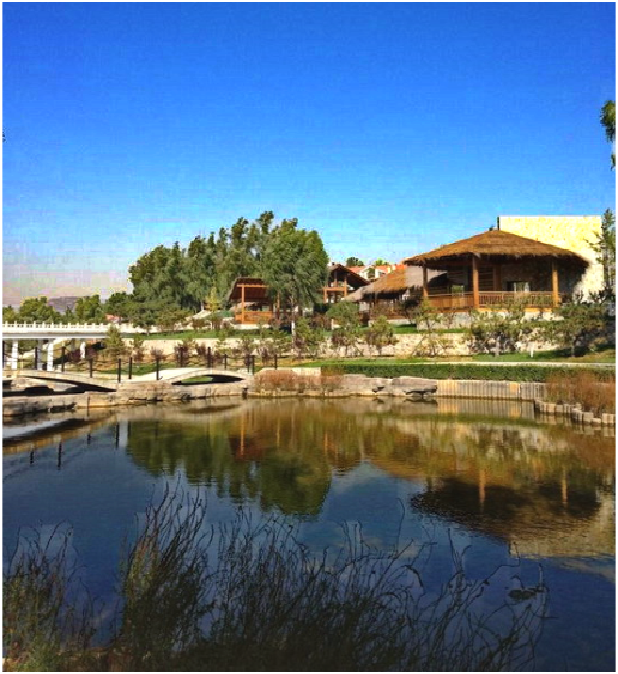} &
    \includegraphics[width=0.18\textwidth]{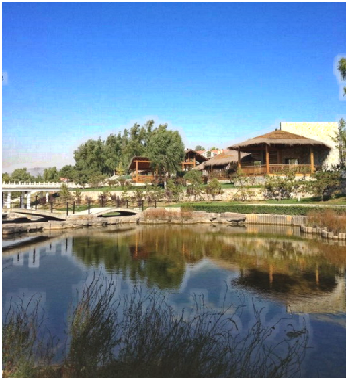} &
    \includegraphics[width=0.18\textwidth]{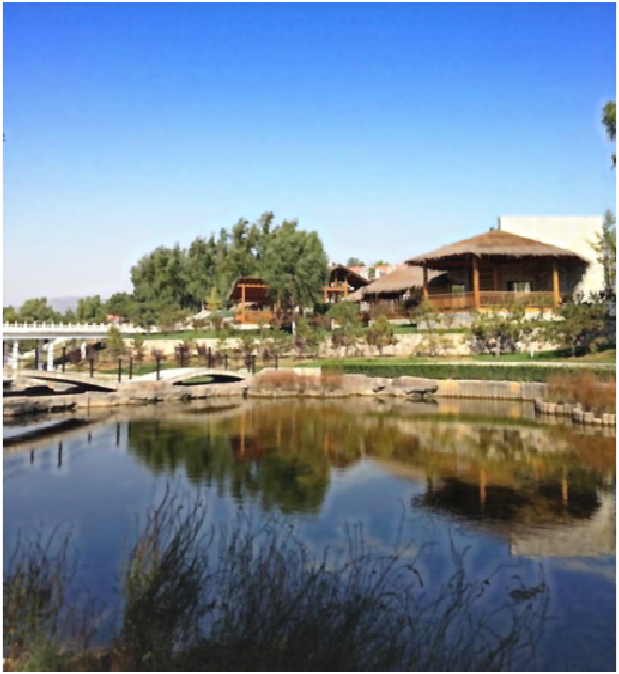} \\[-2pt]

    {\footnotesize Foggy (20\%)} &
    {\footnotesize DCP} &
    {\footnotesize MDCP} &
    {\footnotesize SIDVBM} &
    {\footnotesize Proposed} \\[6pt]

    \includegraphics[width=0.18\textwidth]{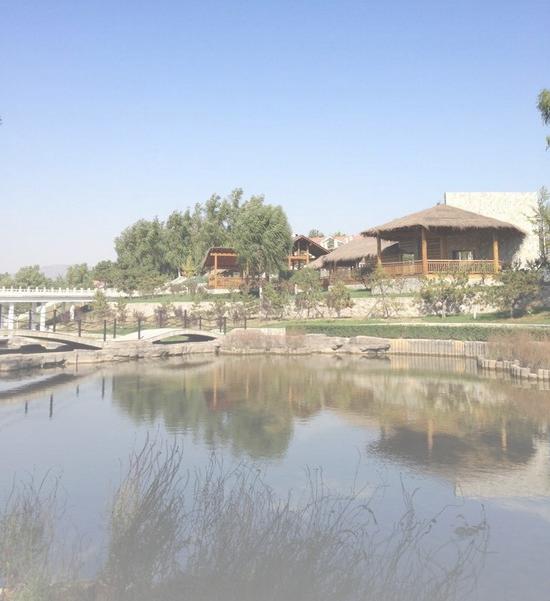} &
    \includegraphics[width=0.18\textwidth]{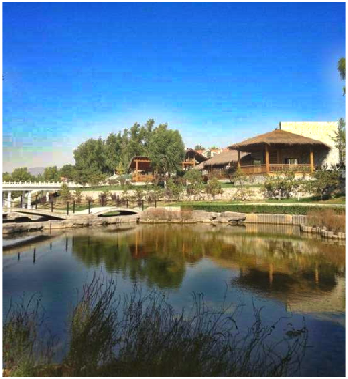} &
    \includegraphics[width=0.18\textwidth]{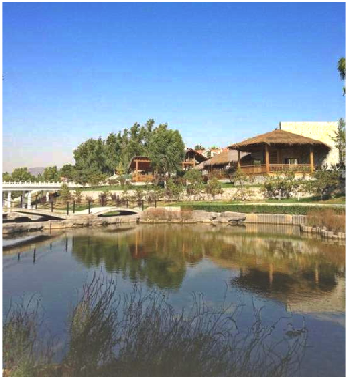} &
    \includegraphics[width=0.18\textwidth]{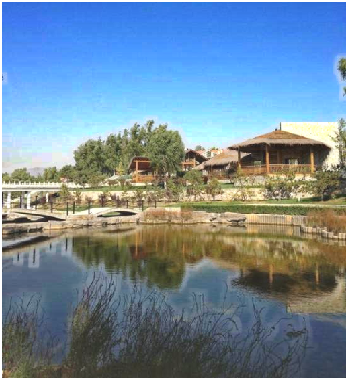} &
    \includegraphics[width=0.18\textwidth]{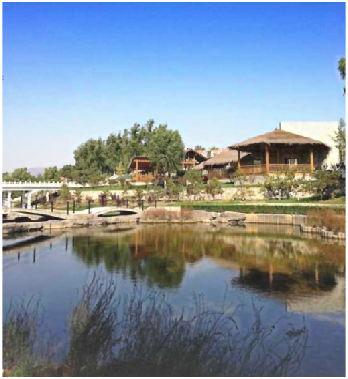} \\[-2pt]

    {\footnotesize Foggy (30\%)} &
    {\footnotesize DCP} &
    {\footnotesize MDCP} &
    {\footnotesize SIDVBM} &
    {\footnotesize Proposed}

    \end{tabular}

    \caption{Visual Comparison of defogging for Img 5. The first column shows the input images, and the second, third, and fourth columns contain the restored versions obtained using DCP\cite{he2011single}, MDCP\cite{salazar2019fast}, SIDVBM\cite{liu2019unified}, and the proposed Model.}

    \label{fig:defogging_comparison_img5}
\end{figure}
\begin{figure}[H]
\centering

\begin{subfigure}{0.48\textwidth}
    \centering
    \includegraphics[width=\linewidth]{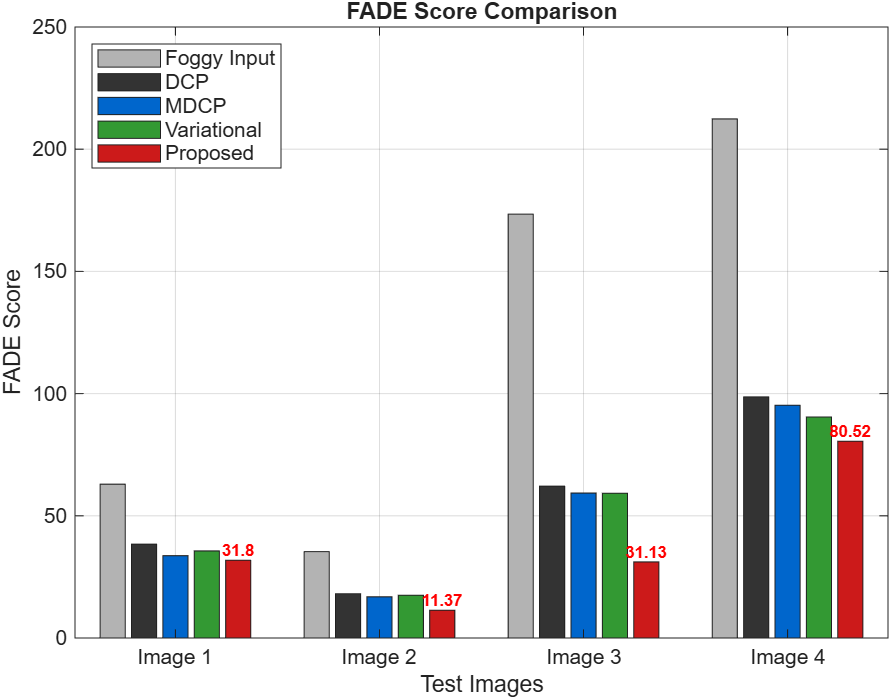}
    \caption{FADE comparison for non-reference images using DCP, MDCP, and the proposed method}
    \label{fig:fade_comparison}
\end{subfigure}
\hfill
\begin{subfigure}{0.48\textwidth}
    \centering
    \includegraphics[width=\linewidth]{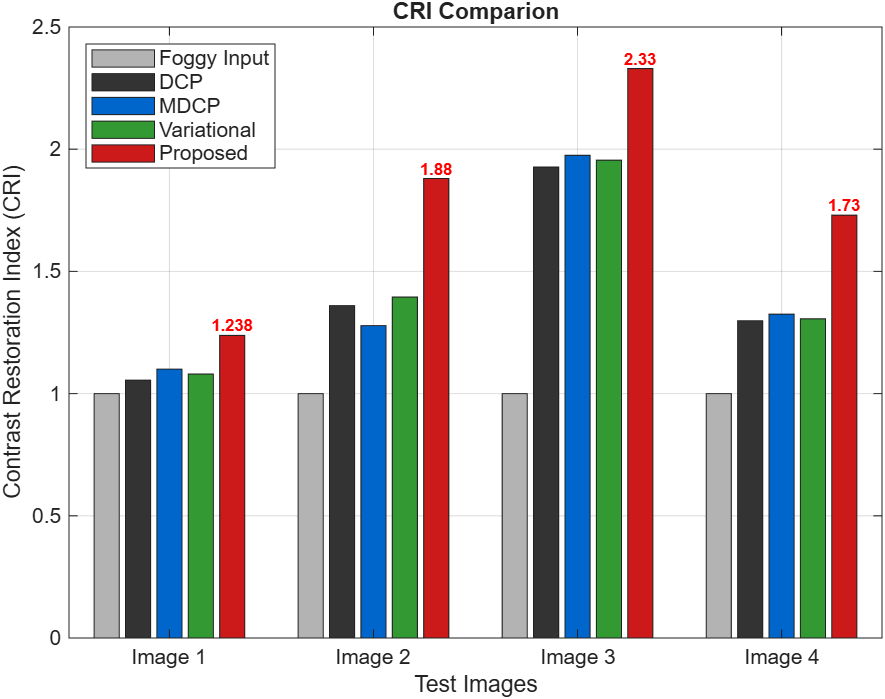}
    \caption{CRI comparison for non-reference images using DCP, MDCP, and the proposed method}
    \label{fig:cri_comparison}
\end{subfigure}

\caption{Comparison of no-reference quality metrics}
\label{fig:fade_cri_comparison}
\end{figure}

\begin{figure}[H]
\centering

\begin{subfigure}{0.48\textwidth}
    \centering
    \includegraphics[width=\linewidth]{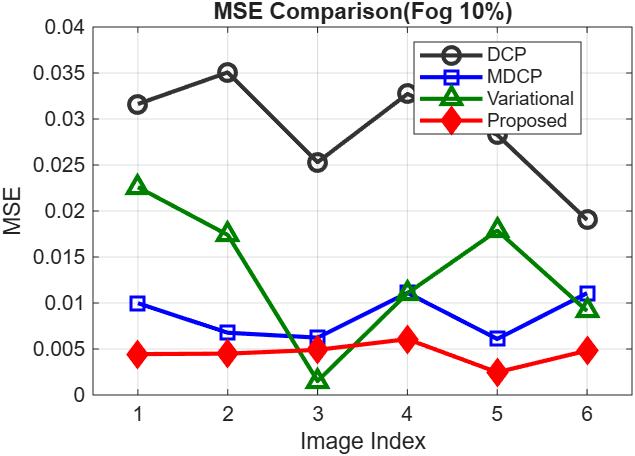}
    
\end{subfigure}
\hfill
\begin{subfigure}{0.48\textwidth}
    \centering
    \includegraphics[width=\linewidth]{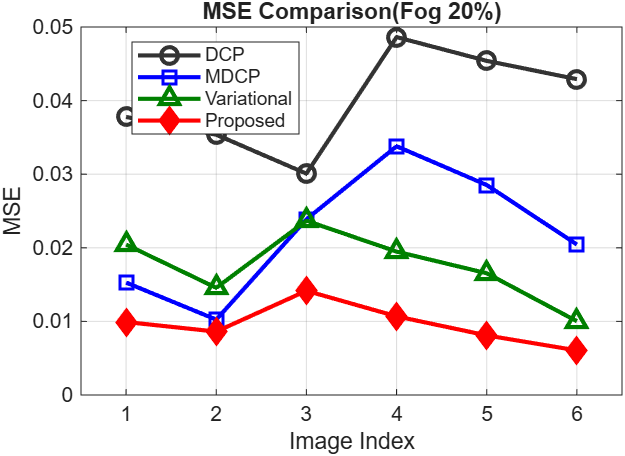}
    
\end{subfigure}

\vspace{0.3cm}

\begin{subfigure}{0.48\textwidth}
    \centering
    \includegraphics[width=\linewidth]{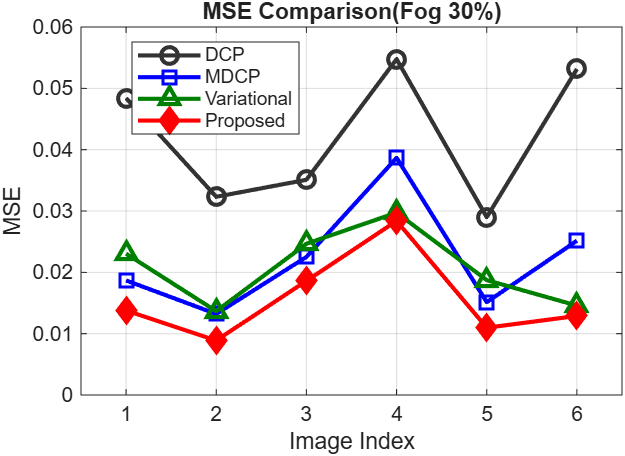}
   
\end{subfigure}
\hfill
\begin{subfigure}{0.48\textwidth}
    \centering
    \includegraphics[width=\linewidth]{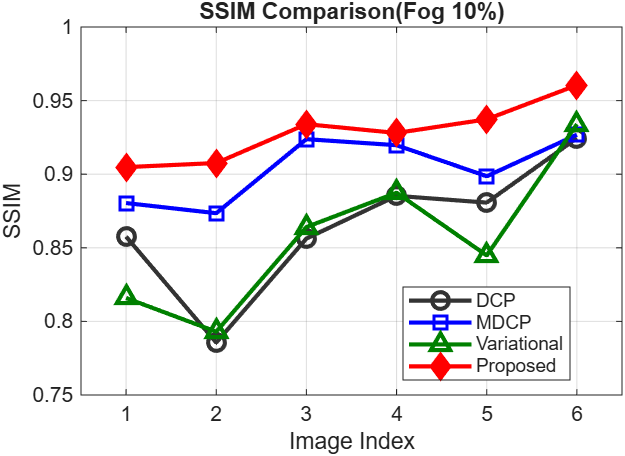}
    
\end{subfigure}

\vspace{0.3cm}

\begin{subfigure}{0.48\textwidth}
    \centering
    \includegraphics[width=\linewidth]{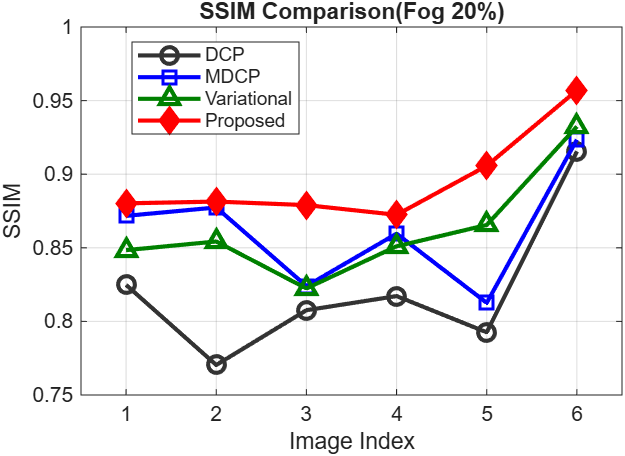}
   
\end{subfigure}
\hfill
\begin{subfigure}{0.48\textwidth}
    \centering
    \includegraphics[width=\linewidth]{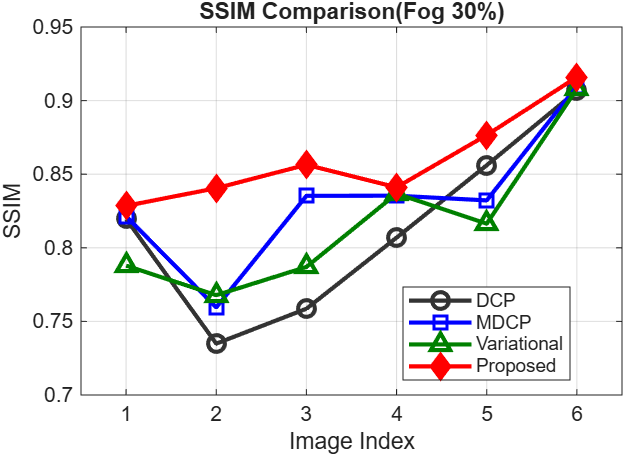}
   
\end{subfigure}

\caption{Line graph comparison of MSE and SSIM of  defogging results obtained using different methods for six images.}
\label{fig:visual_results}
\end{figure}

As observed from the figure ~\ref{fig:defogging_comparison_WGT}, the original foggy images suffer from fog results as lower visibility, low contrast, faded colors due to the presence of dense atmospheric fog. The DCP and MDCP methods shows good improvement; but, some fog artifacts, color distortion, and loss of fine details are still noticeable, especially in regions with complex textures. In contrast, the proposed fourth-order PDE-based defogging method generates visually clearer images in comparison of existing DCP, MDCP, and SIDVBM method. By Figure~\ref{fig:fade_cri_comparison} we presents a visual comparison of two important non reference image quality metrics, namely FADE and CRI, evaluated on four test images for four different defogging methods: DCP, MDCP, SIDVBM and the proposed fourth-order PDE-based approach. by Fig.~\ref{fig:fade_comparison} we see that proposed model give lower value of FADE than DCP, MDCP and SIBVBM methods. The noticeable reduction in FADE across all images confirms that the proposed method is highly capable of suppressing fog and haze effects under different scene conditions rather than DCP, MDCP and SIBVBM methods.
In parallel, the CRI results in Fig.~\ref{fig:cri_comparison} ensure that the ability of each method to restore natural colors after defogging. It can be observed that the proposed method attains the highest CRI values among existing  DCP, MDCP and SIBVBM methods for every test image.
figure~\ref{fig:visual_results} presents the quantitative comparison of defogging performance using MSE and SSIM under different fog densities. The MSE plots at 10, 20, and 30 percent fog levels show that the proposed method produces lower MSE values compared to DCP, MDCP, and SIDVBM methods indicating better fidelity to the reference images. The SSIM comparisons demonstrate that the proposed approach achieves higher structural similarity across all fog conditions, reflecting improved preservation of edges and image details. Overall, the results in Fig.~\ref{fig:visual_results} confirm the robustness of the proposed method in low-density fog and under moderate and dense fog, it achieving good-quality defogging results compared with DCP and MDCP methods.

\begin{table}[H]
\centering
\caption{Performance analysis of image defogging results under three different fog strengths. 
Restoration quality is evaluated using MSE and SSIM, the proposed Model is compared with DCP, MDCP, and the SIDVBM models.}
\label{tab:defogging_mse_ssim}

\setlength{\tabcolsep}{4pt}
\renewcommand{\arraystretch}{1.2}

\begin{tabular}{|c|c|cc|cc|cc|cc|}
\hline
\multirow{2}{*}{Image} & \multirow{2}{*}{Fog Level}
& \multicolumn{2}{c|}{DCP}
& \multicolumn{2}{c|}{MDCP}
& \multicolumn{2}{c|}{Variational}
& \multicolumn{2}{c|}{Proposed} \\ \cline{3-10}

 &  & MSE & SSIM
    & MSE & SSIM
    & MSE & SSIM
    & \textbf{MSE} & \textbf{SSIM} \\ \hline

\multirow{3}{*}{Img 1}
& 10\% & 0.031630 & 0.8576 & 0.009974 & 0.8804 & 0.022589 & 0.8159 & \textbf{0.004445} & \textbf{0.9048} \\
& 20\% & 0.037782 & 0.8248 & 0.015276 & 0.8717 & 0.020443 & 0.8484 & \textbf{0.009917} & \textbf{0.8802} \\
& 30\% & 0.048411 & 0.8196 & 0.018674 & 0.8214 & 0.023035 & 0.7879 & \textbf{0.013769} & \textbf{0.8285} \\ \hline

\multirow{3}{*}{Img 2}
& 10\% & 0.035066 & 0.7856 & 0.006766 & 0.8735 & 0.017398 & 0.7930 & \textbf{0.004496} & \textbf{0.9076} \\
& 20\% & 0.035370 & 0.7704 & 0.010242 & 0.8774 & 0.014565 & 0.8543 & \textbf{0.008614} & \textbf{0.8814} \\
& 30\% & 0.032319 & 0.7349 & 0.013214 & 0.7595 & 0.013667 & 0.7678 & \textbf{0.008890} & \textbf{0.8405} \\ \hline

\multirow{3}{*}{Img 3}
& 10\% & 0.025304 & 0.8563 & 0.006234 & 0.9238 & 0.001519 & 0.8641 & \textbf{0.004906} & \textbf{0.9340} \\
& 20\% & 0.030079 & 0.8076 & 0.023872 & 0.8240 & 0.023681 & 0.8224 & \textbf{0.014157} & \textbf{0.8790} \\
& 30\% & 0.035085 & 0.7586 & 0.022509 & 0.8353 & 0.024659 & 0.7870 & \textbf{0.018606} & \textbf{0.8564} \\ \hline

\multirow{3}{*}{Img 4}
& 10\% & 0.032725 & 0.8853 & 0.011101 & 0.9197 & 0.011061 & 0.8874 & \textbf{0.006073} & \textbf{0.9280} \\
& 20\% & 0.048625 & 0.8172 & 0.033796 & 0.8596 & 0.019482 & 0.8510 & \textbf{0.010686} & \textbf{0.8724} \\
& 30\% & 0.054759 & 0.8069 & 0.038670 & 0.8354 & 0.029727 & 0.8368 & \textbf{0.028379} & \textbf{0.8411} \\ \hline

\multirow{3}{*}{Img 5}
& 10\% & 0.028280 & 0.8807 & 0.006090 & 0.8986 & 0.017874 & 0.8448 & \textbf{0.002468} & \textbf{0.9373} \\
& 20\% & 0.045400 & 0.7926 & 0.028529 & 0.8126 & 0.016543 & 0.8655 & \textbf{0.008097} & \textbf{0.9057} \\
& 30\% & 0.028986 & 0.8561 & 0.015136 & 0.8322 & 0.018723 & 0.8164 & \textbf{0.010972} & \textbf{0.8766} \\ \hline

\multirow{3}{*}{Img 6}
& 10\% & 0.019024 & 0.9245 & 0.011073 & 0.9268 & 0.009132 & 0.9340 & \textbf{0.004823} & \textbf{0.9605} \\
& 20\% & 0.042914 & 0.9155 & 0.020445 & 0.9235 & 0.010023 & 0.9324 & \textbf{0.006054} & \textbf{0.9570} \\
& 30\% & 0.053155 & 0.9066 & 0.025153 & 0.9098 & 0.014576 & 0.9080 & \textbf{0.012887} & \textbf{0.9160} \\ \hline

\end{tabular}
\end{table}

\begin{table}[H]
\centering
\caption{Comparative analysis of FADE, CRI, Entropy, and Average Gradient (AG) metrics for foggy images processed by DCP, MDCP, SIDVBM, and the proposed model across non reference color images.}
\label{tab:defogging_qualitative_metrics}

\setlength{\tabcolsep}{5pt}
\renewcommand{\arraystretch}{1.2}

\begin{tabular}{|c|c|c|c|c|c|}
\hline
Image & Method & FADE & CRI & Entropy & AG \\ \hline

\multirow{5}{*}{Image 1}
& Foggy        & 62.93 & 1.000 & 7.08 & 0.0103 \\
& DCP          & 38.38 & 1.055 & 7.53 & 0.0144 \\
& MDCP         & 33.68 & 1.100 & 7.55 & 0.0161 \\
& SIDVBM  & 35.62 & 1.08 & 7.57 & 0.0158 \\
& \textbf{Proposed} & \textbf{31.80} & \textbf{1.238} & \textbf{7.67} & \textbf{0.0170} \\ \hline

\multirow{5}{*}{Image 2}
& Foggy        & 35.33 & 1.000 & 6.43 & 0.0171 \\
& DCP          & 18.11 & 1.360 & 6.65 & 0.0250 \\
& MDCP         & 16.85 & 1.278 & 6.48 & 0.0263 \\
& SIDVBM  & 17.48 & 1.395 & 6.66 & 0.0266 \\
& \textbf{Proposed} & \textbf{11.37} & \textbf{1.88} & \textbf{6.84} & \textbf{0.0326} \\ \hline

\multirow{5}{*}{Image 3}
& Foggy        & 173.44 & 1.000 & 5.75 & 0.0004 \\
& DCP          & 62.12  & 1.927  & 6.67 & 0.0092 \\
& MDCP         & 59.31  & 1.975  & 6.42 & 0.0095 \\
& SIDVBM  & 59.24  & 1.955 & 6.52 & 0.0098 \\
& \textbf{Proposed} & \textbf{31.13} & \textbf{2.33} & \textbf{7.34} & \textbf{0.0137} \\ \hline

\multirow{5}{*}{Image 4}
& Foggy        & 212.39 & 1.000 & 6.63 & 0.0031 \\
& DCP          & 98.67  & 1.298 & 6.73 & 0.0050 \\
& MDCP         & 95.24  & 1.325 & 6.06 & 0.0051 \\
& SIDVBM  & 90.42  & 1.306 & \textbf{6.95} & \textbf{0.0055} \\
& \textbf{Proposed} & \textbf{80.52} & \textbf{1.73} & 6.89 & 0.0065 \\ \hline
\end{tabular}
\end{table}

\begin{figure}[H]
    \centering
    \setlength{\tabcolsep}{2pt}
    \renewcommand{\arraystretch}{1.0}
    \begin{tabular}{ccccc}

        \includegraphics[width=0.19\textwidth]{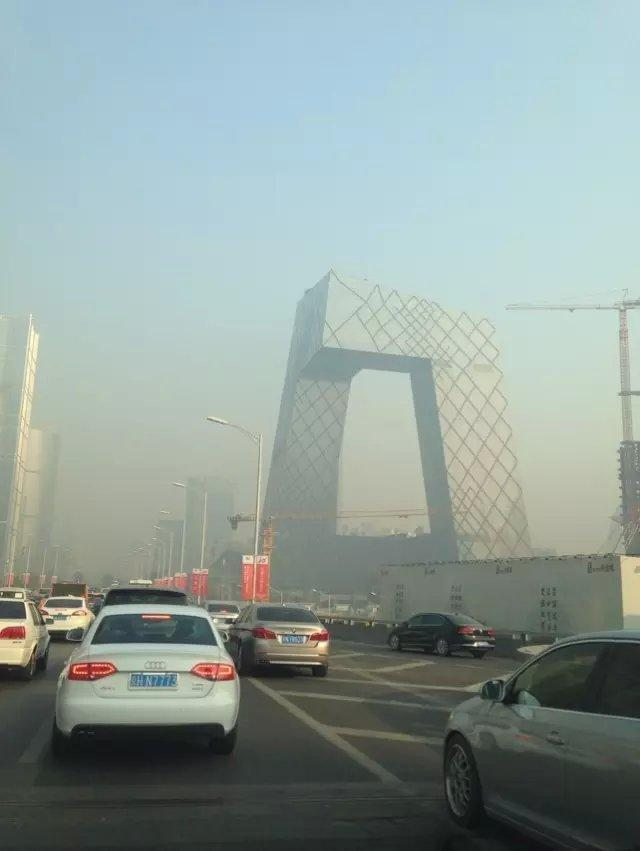} &
        \includegraphics[width=0.19\textwidth]{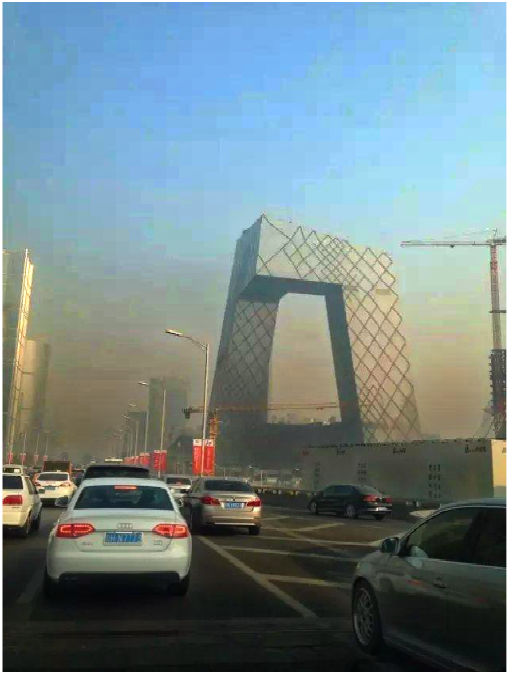} &
        \includegraphics[width=0.19\textwidth]{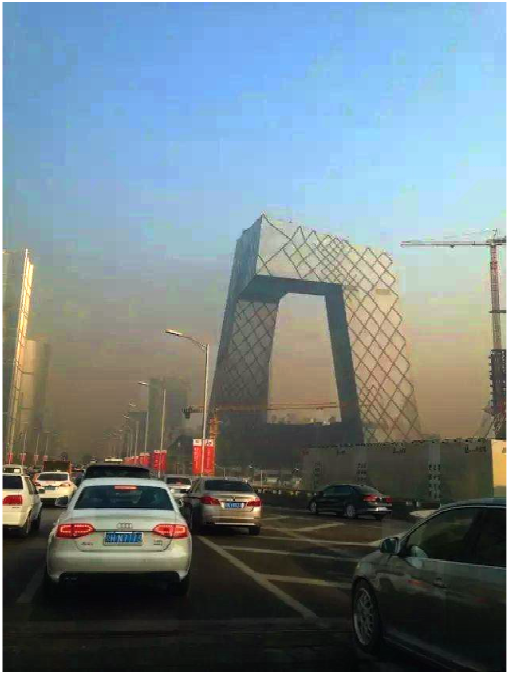} &
        \includegraphics[width=0.19\textwidth]{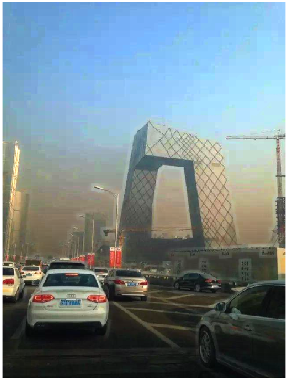} &
        \includegraphics[width=0.19\textwidth]{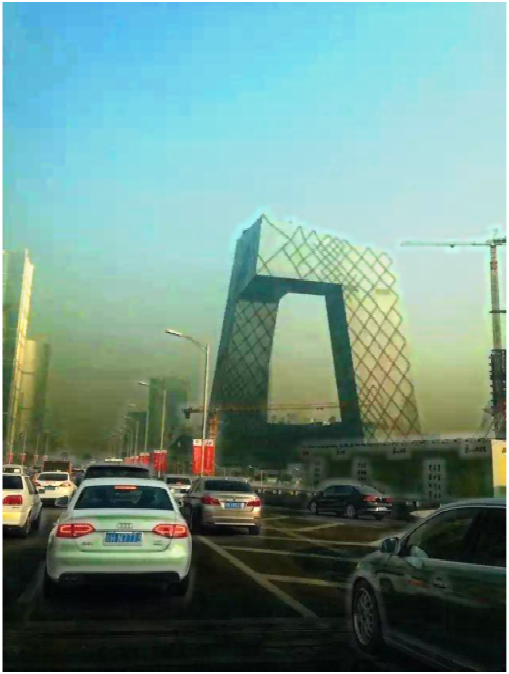} \\

        \includegraphics[width=0.19\textwidth]{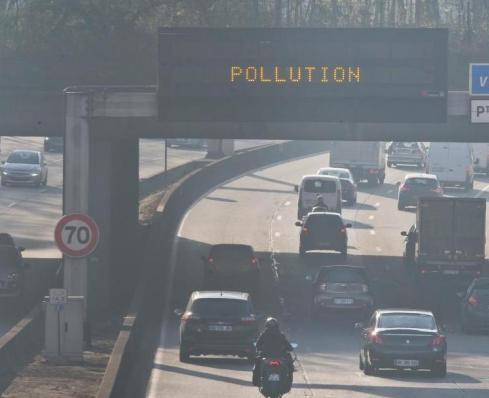} &
        \includegraphics[width=0.19\textwidth]{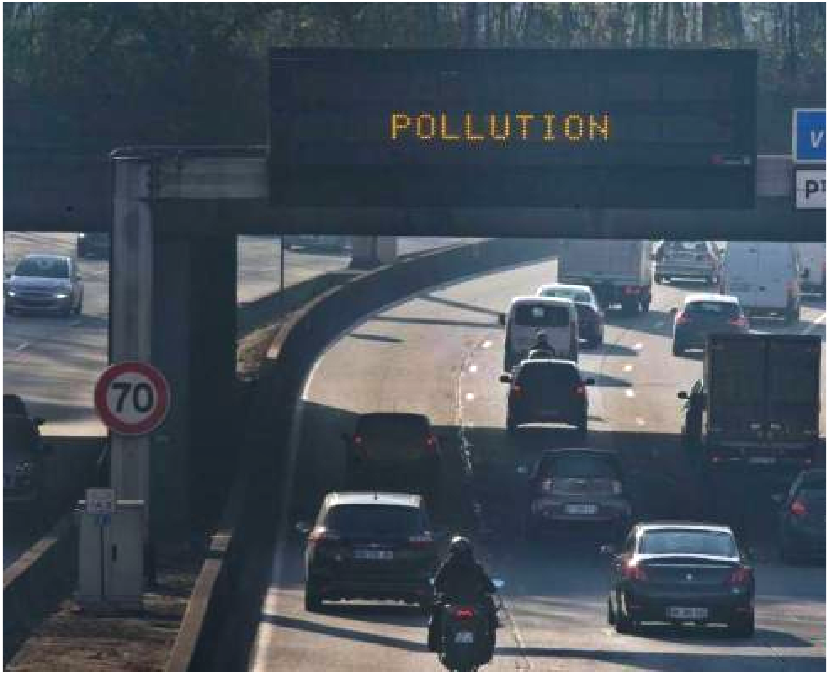} &
        \includegraphics[width=0.19\textwidth]{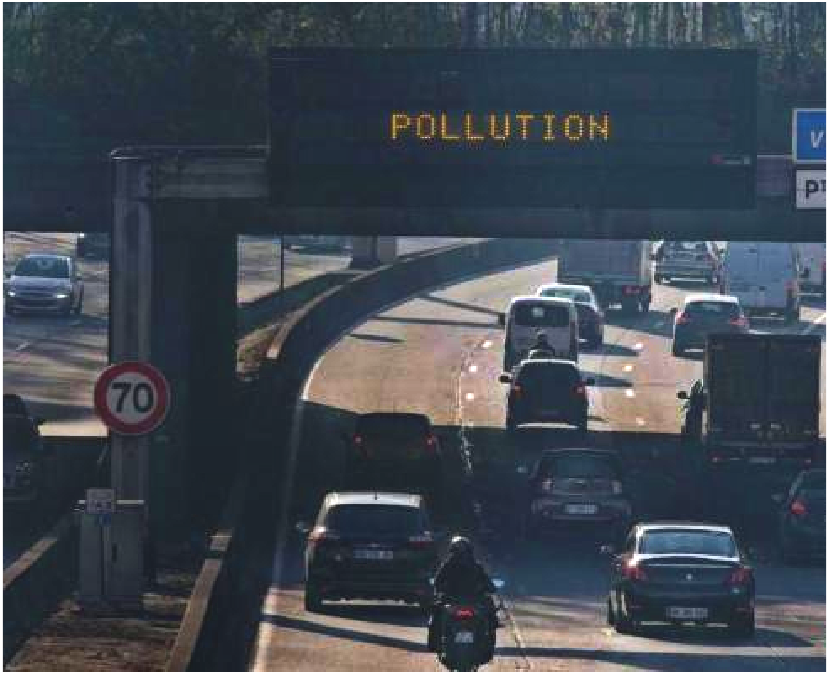} &
        \includegraphics[width=0.19\textwidth]{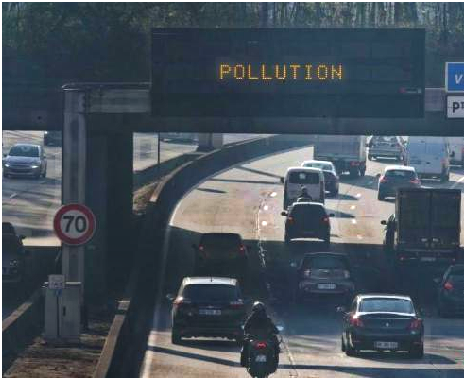} &
        \includegraphics[width=0.19\textwidth]{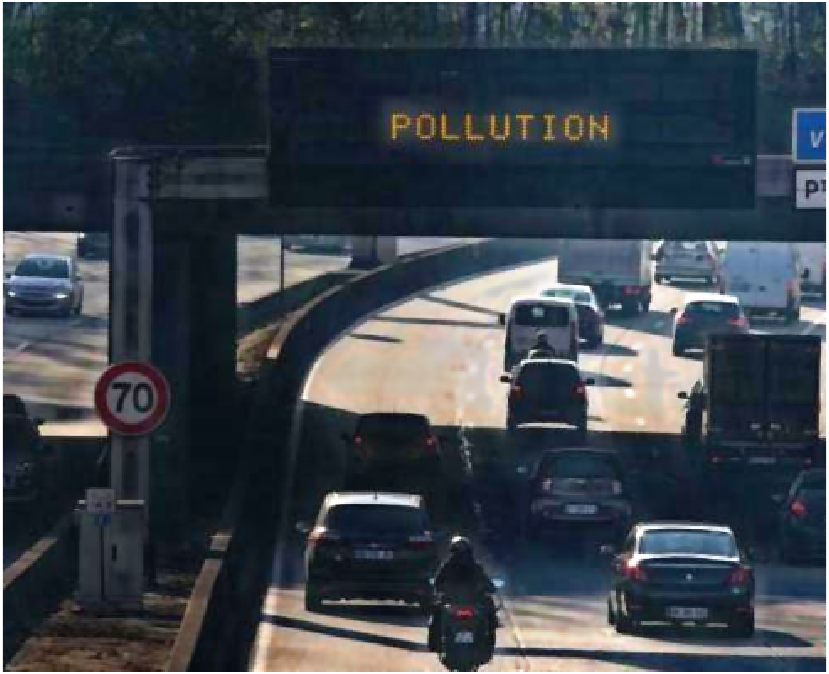} \\

        \includegraphics[width=0.19\textwidth]{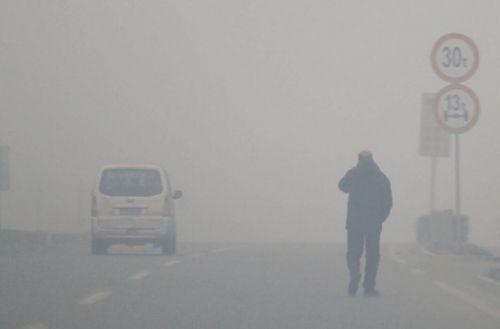} &
        \includegraphics[width=0.19\textwidth]{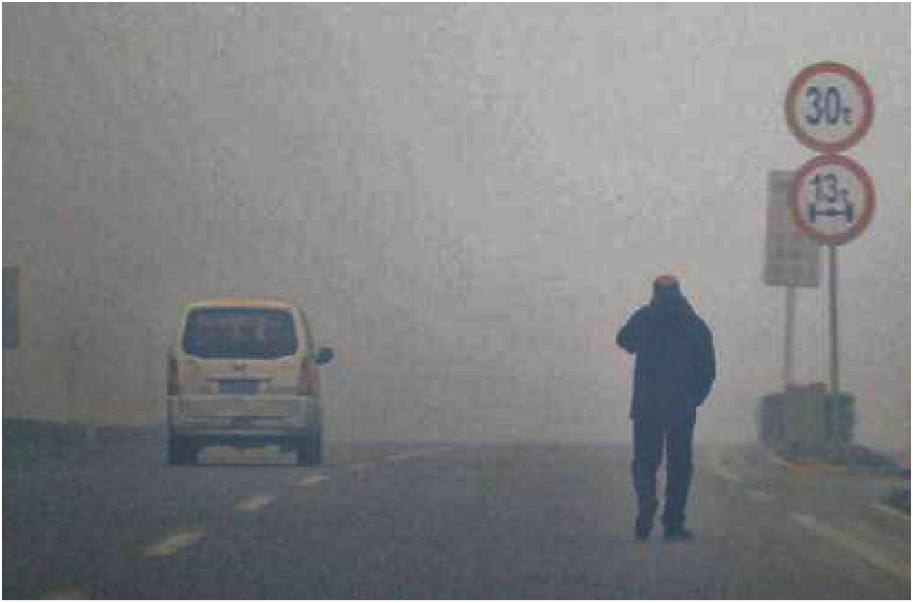} &
        \includegraphics[width=0.19\textwidth]{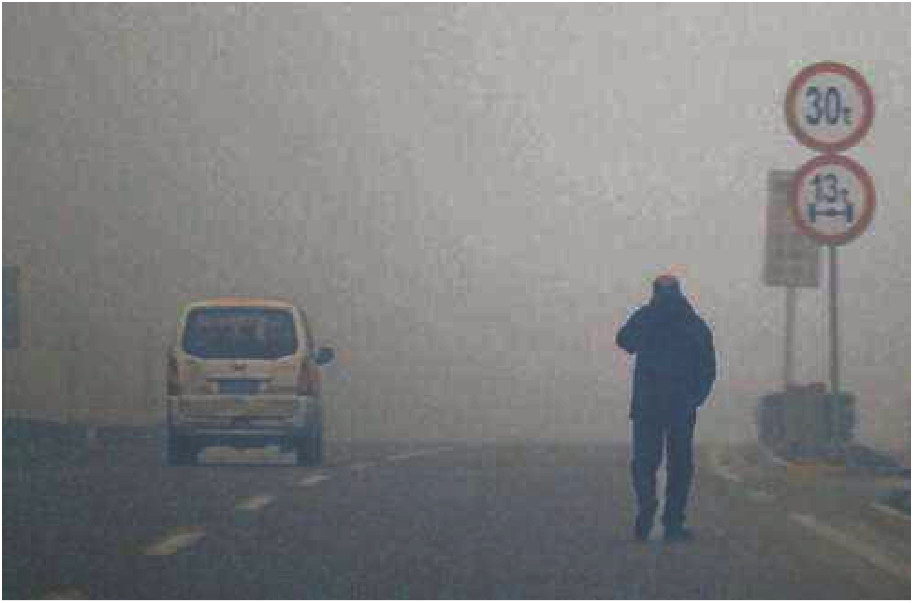} &
        \includegraphics[width=0.19\textwidth]{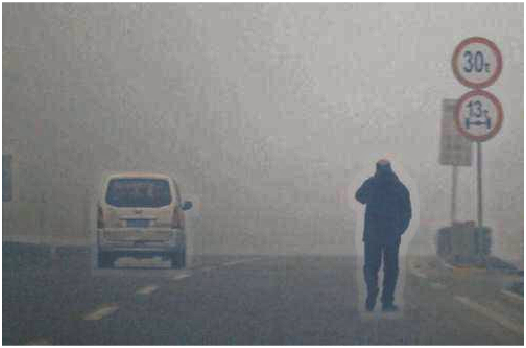} &
        \includegraphics[width=0.19\textwidth]{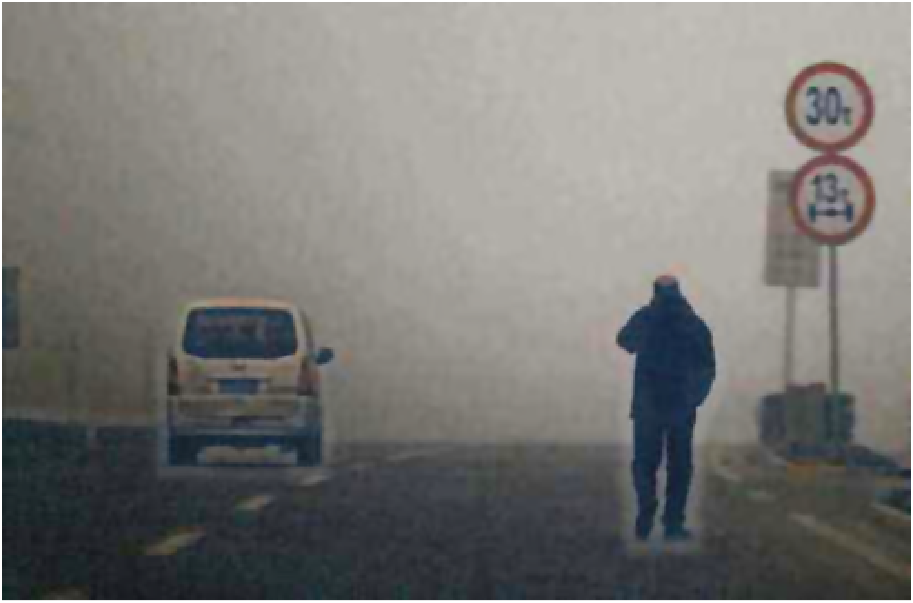} \\

        \includegraphics[width=0.19\textwidth]{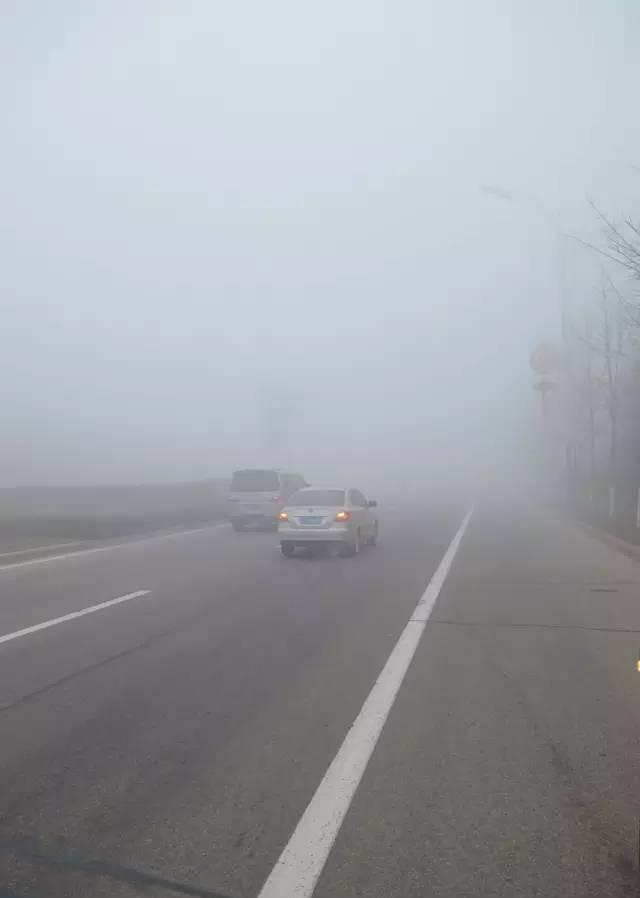} &
        \includegraphics[width=0.19\textwidth]{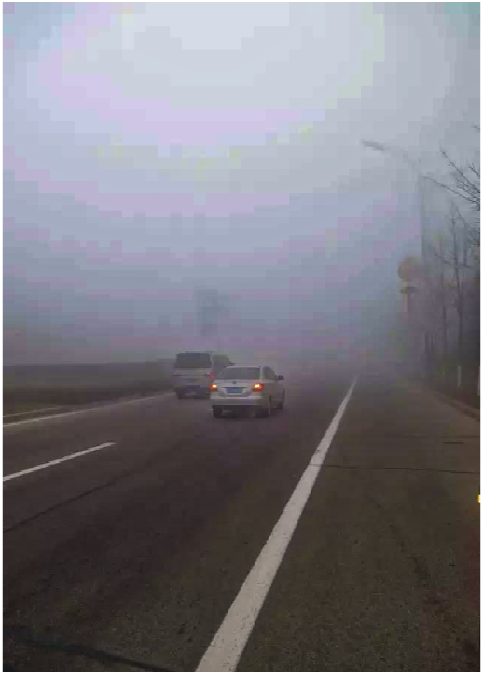} &
        \includegraphics[width=0.19\textwidth]{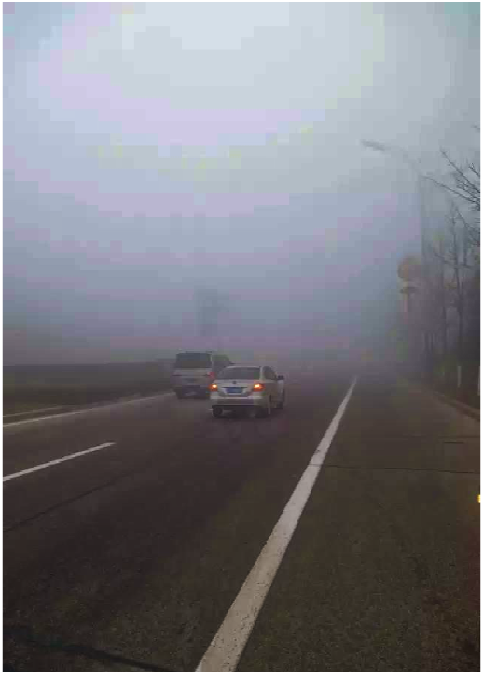} &
        \includegraphics[width=0.19\textwidth]{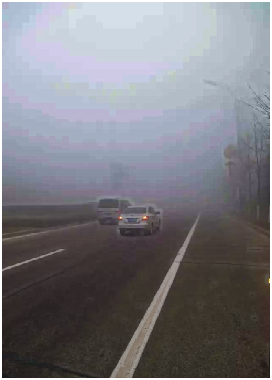} &
        \includegraphics[width=0.19\textwidth]{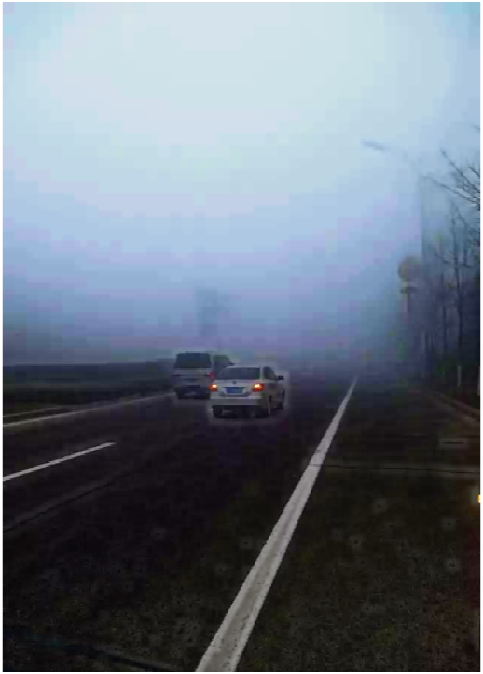} \\

        \small (a) Foggy &
        \small (b) DCP &
        \small (c) MDCP &
        \small (d) Variational &
        \small (e) Proposed \\

    \end{tabular}

    \caption{The first column contains input foggy non-references image, second third and forth column contains de-hazed images using Dark Channel Prior (DCP),
    Modified Dark Channel Prior (MDCP), SIDVBM, and the proposed Model.}
    \label{fig:defogging_comparison_WGT}
\end{figure}

\section{Conclusion and Future Work}

This paper propose a fourth-order PDE-based image defogging model used to defogging of color images that are degraded by atmospheric fog. Distinct from traditional defogging techniques, the proposed method applies higher-order diffusion to suppress fog while preserving structural details. The model adapts well to different densities, making it suitable for a wide range of outdoor scenes.
The performance was evaluated using both full-reference and no-reference image quality metrics. The quantitative analysis using MSE and SSIM demonstrates that the proposed approach achieves lower reconstruction error and higher structural similarity when compared with DCP. MDCP, and SIDVBM methods. This shows that the proposed model preserve more image structural details while reducing fog. For the no-reference images quality evaluation we using FADE, CRI, entropy, and average gradient.
For future work, several directions remain open for research. The framework can be extended to video defogging. AND, adaptive parameter selection and learning-based strategies may be explored to further improve robustness under highly varying atmospheric conditions.


\begin{thebibliography}{99}

\bibitem{koschmieder1924theorie}
H.~Koschmieder, ``Theorie der horizontalen Sichtweite,'' \textit{Beiträge zur Physik der freien Atmosphäre}, vol.~12, pp.~33--53, 1924.

\bibitem{mccartney1976optics}
E.~J. McCartney, \textit{Optics of the Atmosphere: Scattering by Molecules and Particles}. New York, NY, USA: John Wiley and Sons, 1976.

\bibitem{narasimhan2002vision}
S.~G. Narasimhan and S.~K. Nayar, ``Vision and the atmosphere,'' \textit{International Journal of Computer Vision}, vol.~48, no.~3, pp.~233--254, 2002.

\bibitem{narasimhan2003contrast}
S.~G. Narasimhan and S.~K. Nayar, ``Contrast restoration of weather degraded images,'' \textit{IEEE Transactions on Pattern Analysis and Machine Intelligence}, vol.~25, no.~6, pp.~713--724, 2003.

\bibitem{tan2008visibility}
R.~Tan, ``Visibility in bad weather from a single image,'' in \textit{IEEE Conference on Computer Vision and Pattern Recognition (CVPR)}, 2008, pp.~1--8.

\bibitem{fattal2008single}
R.~Fattal, ``Single image dehazing,'' \textit{ACM Transactions on Graphics (SIGGRAPH)}, vol.~27, no.~3, article~72, 2008.

\bibitem{tarel2009fast}
J.~P. Tarel and N.~Hautiere, ``Fast visibility restoration from a single color or gray level image,'' in \textit{IEEE International Conference on Computer Vision (ICCV)}, 2009, pp.~2201--2208.

\bibitem{he2009single}
K.~He, J.~Sun, and X.~Tang, ``Single image haze removal using dark channel prior,'' in \textit{IEEE Conference on Computer Vision and Pattern Recognition (CVPR)}, 2009, pp.~1956--1963.

\bibitem{he2011single}
K.~He, J.~Sun, and X.~Tang, ``Single image haze removal using dark channel prior,'' \textit{IEEE Transactions on Pattern Analysis and Machine Intelligence}, vol.~33, no.~12, pp.~2341--2353, 2011.

\bibitem{he2013guided}
K.~He, J.~Sun, and X.~Tang, ``Guided image filtering,'' \textit{IEEE Transactions on Pattern Analysis and Machine Intelligence}, vol.~35, no.~6, pp.~1397--1409, 2013.

\bibitem{tufail2018improved}
Z.~Tufail, K.~Khurshid, A.~Salman, I.~F. Nizami, K.~Khurshid, and B.~Jeon, ``Improved Dark Channel Prior for Image Defogging Using RGB and YCbCr Color Space,'' \textit{IEEE Access}, vol.~6, pp.~32576--32587, 2018.

\bibitem{sabir2020segmentation}
A.~Sabir, K.~Khurshid, and A.~Salman, ``Segmentation-based image defogging using modified dark channel prior,'' \textit{EURASIP Journal on Image and Video Processing}, vol.~2020, no.~6, 2020.

\bibitem{anan2021image}
S.~Anan, M.~I. Khan, M.~M.~S. Kowsar, K.~Deb, P.~K. Dhar, and T.~Koshiba, ``Image Defogging Framework Using Segmentation and the Dark Channel Prior,'' \textit{Entropy}, vol.~23, no.~3, p.~285, 2021.

\bibitem{wu2024real}
X.~Wu, X.~Chen, X.~Wang, X.~Zhang, S.~Yuan, B.~Sun, X.~Huang, and L.~Liu, ``A real-time framework for HD video defogging using modified dark channel prior,'' \textit{Journal of Real-Time Image Processing}, vol.~21, no.~55, 2024.

\bibitem{kokul2020single}
T.~Kokul and S.~Anparasy, ``Single Image Defogging using Depth Estimation and Scene-Specific Dark Channel Prior,'' in \textit{20th International Conference on Advances in ICT for Emerging Regions (ICTer)}, 2020, pp.~190--195.

\bibitem{li2017improved}
C.~Li, T.~Fan, X.~Ma, Z.~Zhang, H.~Wu, and L.~Chen, ``An Improved Image Defogging Method Based on Dark Channel Prior,'' in \textit{2017 2nd International Conference on Image, Vision and Computing}, 2017, pp.~1--5.

\bibitem{chen2009single}
M.~Chen, A.~Men, P.~Fan, and B.~Yang, ``Single image defogging,'' in \textit{Proceedings of IC-NIDC}, 2009, pp.~1--5.

\bibitem{gibson2013analysis}
K.~B. Gibson and T.~Q. Nguyen, ``An analysis of single image defogging methods using a color ellipsoid framework,'' \textit{EURASIP Journal on Image and Video Processing}, vol.~2013, no.~37, 2013.

\bibitem{pandey2025comprehensive}
P.~Pandey, R.~Gupta, and N.~Goel, ``Comprehensive review of single image defogging techniques: enhancement, prior, and learning based approaches,'' \textit{Artificial Intelligence Review}, vol.~58, p.~116, 2025.

\bibitem{zhu2015fast}
Q.~Zhu, J.~Mai, and L.~Shao, ``A fast single image haze removal algorithm using color attenuation prior,'' \textit{IEEE Transactions on Image Processing}, vol.~24, no.~11, pp.~3522--3533, 2015.

\bibitem{meng2013efficient}
G.~Meng, Y.~Wang, J.~Duan, S.~Xiang, and C.~Pan, ``Efficient image dehazing with boundary constraint and contextual regularization,'' in \textit{IEEE International Conference on Computer Vision (ICCV)}, 2013, pp.~617--624.

\bibitem{berman2016non}
D.~Berman, S.~Avidan, \textit{et al.}, ``Non-local image dehazing,'' in \textit{IEEE Conference on Computer Vision and Pattern Recognition (CVPR)}, 2016, pp.~1674--1682.

\bibitem{nishino2012bayesian}
K.~Nishino, L.~Kratz, and S.~Lombardi, ``Bayesian defogging,'' \textit{International Journal of Computer Vision}, vol.~98, no.~3, pp.~263--278, 2012.

\bibitem{levin2008closed}
A.~Levin, D.~Lischinski, and Y.~Weiss, ``A closed-form solution to natural image matting,'' \textit{IEEE Transactions on Pattern Analysis and Machine Intelligence}, vol.~30, no.~2, pp.~228--242, 2008.

\bibitem{ancuti2013single}
C.~O. Ancuti and C.~Ancuti, ``Single image dehazing by multi-scale fusion,'' \textit{IEEE Transactions on Image Processing}, vol.~22, no.~8, pp.~3271--3282, 2013.

\bibitem{cai2016dehazenet}
B.~Cai, X.~Xu, K.~Jia, C.~Chun, and D.~Tao, ``DehazeNet: An end-to-end system for single image haze removal,'' \textit{IEEE Transactions on Image Processing}, vol.~25, no.~11, pp.~5187--5198, 2016.

\bibitem{ren2016single}
W.~Ren, S.~Liu, H.~Zhang, J.~Pan, X.~Cao, and M.-H. Yang, ``Single image dehazing via multi-scale convolutional neural networks,'' in \textit{European Conference on Computer Vision (ECCV)}, 2016, pp.~154--169.

\bibitem{li2017aod}
B.~Li, X.~Peng, Z.~Wang, J.~Xu, and D.~Feng, ``AOD-Net: All-in-one dehazing network,'' in \textit{IEEE International Conference on Computer Vision (ICCV)}, 2017, pp.~4770--4778.

\bibitem{zhang2018densely}
H.~Zhang and V.~M. Patel, ``Densely connected pyramid dehazing network,'' in \textit{IEEE Conference on Computer Vision and Pattern Recognition (CVPR)}, 2018, pp.~8193--8202.

\bibitem{zhu2018haze}
Y.~Zhu, G.~Tang, X.~Zhang, J.~Jiang, and Q.~Tian, ``Haze removal method for natural restoration of images with sky,'' \textit{Neurocomputing}, vol.~275, pp.~499--510, 2018.

\bibitem{huang2014efficient}
S.~C. Huang, B.~H. Chen, and Y.~J. Cheng, ``An Efficient Visibility Enhancement Algorithm for Road Scenes Captured by Intelligent Transportation Systems,'' \textit{IEEE Transactions on Intelligent Transportation Systems}, vol.~15, no.~5, pp.~2321--2332, 2014.

\bibitem{salazar2019fast}
S. Salazar-Colores, E. Cabal-Y{\'e}pez, J. M. Ramos-Arregu{\'\i}n, 
G. Botella, L. M. Ledesma-Carrillo, and S. Ledesma,
A fast image dehazing algorithm using morphological reconstruction,
IEEE Transactions on Image Processing, vol. 28, no. 5, pp. 2357--2366, 2019.

\bibitem{tarel2012vision}
J.-P. Tarel, N.~Hautiere, L.~Caraffa, A.~Cord, H.~Halmaoui, and D.~Gruyer, ``Vision enhancement in homogeneous and heterogeneous fog,'' \textit{IEEE Intelligent Transportation Systems Magazine}, vol.~4, no.~2, pp.~6--20, 2012.

\bibitem{land1977retinex}
E.~H. Land, ``The retinex theory of color vision,'' \textit{Scientific American}, vol.~237, no.~6, pp.~108--128, 1977.

\bibitem{pizer1987adaptive}
S.~M. Pizer \textit{et al.}, ``Adaptive histogram equalization and its variations,'' \textit{Computer Vision, Graphics, and Image Processing}, vol.~39, no.~3, pp.~355--368, 1987.

\bibitem{kopf2008deep}
J.~Kopf, B.~Neubert, B.~Chen, M.~Cohen, D.~Cohen-Or, O.~Deussen, and M.~Uyttendaele, ``Deep photo: Model-based photograph enhancement and viewing,'' \textit{ACM Transactions on Graphics}, vol.~27, no.~5, pp.~116:1--116:10, 2008.

\bibitem{schechner2001instant}
Y.~Y. Schechner, S.~G. Narasimhan, and S.~K. Nayar, ``Instant dehazing of images using polarization,'' in \textit{IEEE Computer Society Conference on Computer Vision and Pattern Recognition (CVPR)}, 2001, vol.~1, pp.~325--332.

\bibitem{perona1990scale}
P.~Perona and J.~Malik, ``Scale-space and edge detection using anisotropic diffusion,'' \textit{IEEE Transactions on Pattern Analysis and Machine Intelligence}, vol.~12, no.~7, pp.~629--639, 1990.

\bibitem{rudin1992nonlinear}
L.~Rudin, S.~Osher, and E.~Fatemi, ``Nonlinear total variation based noise removal algorithms,'' \textit{Physica D: Nonlinear Phenomena}, vol.~60, no.~1-4, pp.~259--268, 1992.

\bibitem{chan2005image}
T.~Chan and J.~Shen, \textit{Image Processing and Analysis: Variational, PDE, Wavelet, and Stochastic Methods}. SIAM, 2005.

\bibitem{weickert1998anisotropic}
J.~Weickert, \textit{Anisotropic Diffusion in Image Processing}. Teubner Stuttgart, 1998.

\bibitem{majee2020despeckling} 
S. Majee, R. K. Ray, A. K. Majee, A gray level indicator-based regularized telegraph diffusion model: application to image despeckling, 
\textit{SIAM J. Imaging Sci.}, 13(2) (2020), pp. 844–870.
\bibitem{you2000fourth} 
Y.-L. You, M. Kaveh, Fourth-order partial differential equations for noise removal, 
\textit{IEEE Trans. Image Process.}, 9(10) (2000), pp. 1723–1730.

\bibitem{ray2025new}
R. K. Ray, M. Kumar, New fourth-order grayscale indicator-based telegraph diffusion model for image despeckling,
\textit{arXiv preprint arXiv:2509.26010}, (2025).


\bibitem{li2009numerical}
Z. Li, \textit{Numerical Solutions to Partial Differential Equations}, Higher Education Press, Beijing, 2009.

\bibitem{zauderer2011partial} 
E. Zauderer, \textit{Partial Differential Equations of Applied Mathematics}, Pure Appl. Math. (N. Y.), 71, John Wiley Sons, New York, 2011.
\bibitem{zhang2014tensor} 
W. Zhang, J. Li, Y. Yang, A class of nonlocal tensor telegraph-diffusion equations applied to coherence enhancement, 

\bibitem{fang2014single} 
F. Fang, F. Li, T. Zeng, Single image dehazing and denoising: A fast variational approach, SIAM Journal on Imaging Sciences 7 (2) (2014) 969--996.

\bibitem{galdran2015enhanced} 
A. Galdran, J. Vazquez-Corral, D. Pardo, M. Bertalmio, Enhanced variational image dehazing, SIAM Journal on Imaging Sciences 8 (3) (2015) 1519--1546.

\bibitem{liu2019unified} 
Y. Liu, J. Shang, L. Pan, A. Wang, M. Wang, A unified variational model for single image dehazing, IEEE Access 7 (2019) 2019.
\end{thebibliography}
\end{document}